\documentclass[11pt]{book}
\usepackage[utf8]{inputenc}

\usepackage[margin=1.25in]{geometry}
\usepackage{amsmath, amssymb, amsthm}
\usepackage[english]{babel}
\usepackage{blindtext}
\usepackage{enumerate}
\usepackage{mathtools} 

\usepackage{natbib}
\setcitestyle{authoryear,open={(},close={)}}

\usepackage[nottoc,notlot,notlof]{tocbibind}
\usepackage[titletoc]{appendix}
\usepackage{csquotes} 
\usepackage{bm}

\usepackage{color}
\usepackage[hidelinks]{hyperref} 
\usepackage{tocloft}
\hypersetup{
    colorlinks=false, 
    linktoc=all,     
    linkcolor=blue,  
    citecolor= blue  
}

\usepackage{fancyhdr}
\usepackage{apptools}
\AtAppendix{\counterwithin{lemma}{section}}
\AtAppendix{\counterwithin{definition}{section}}
\AtAppendix{\counterwithin{theorem}{section}}
\AtAppendix{\counterwithin{proposition}{section}}

\newtheorem{theorem}{Theorem}[section]

\newtheorem{definition}[theorem]{Definition}

\newtheorem*{remark}{Remark}
\newtheorem*{definition*}{Definition}
\newtheorem*{lemma*}{Lemma}
\newtheorem*{proposition*}{Proposition}

\theoremstyle{definition}

\newcommand{\labs}{\displaystyle\left\lvert}
\newcommand{\rabs}{\right\rvert}

\newcommand*{\defeq}{\stackrel{\textup{def}}{=}} 
\renewcommand{\&}{\textup{\symbol{`\&}}} 

\DeclareMathOperator*{\argmax}{arg\,max}

\DeclareMathOperator{\sgn}{sgn}
\DeclareMathOperator{\spn}{span}

\title{Literature Review: Graph Kernels in Chemoinformatics}
\author{James Young}
\date{Summer 2022}

\begin{document}

\maketitle


\chapter*{Preface}
\addcontentsline{toc}{chapter}{Preface}

The purpose of this review is to introduce the reader to \emph{graph kernels} and the corresponding literature, with an emphasis on those with direct application to \emph{chemoinformatics}. Graph kernels are functions that allow for the inference of properties of molecules and compounds, which can help with tasks such as finding suitable compounds in drug design. The use of kernel methods is but one particular way two quantify similarity between graphs. We restrict our discussion to this one method, although popular alternatives have emerged in recent years, most notably graph neural networks.

\smallskip

The first two chapters provide the necessary background on statistical learning theory and common techniques. Chapter \ref{chapter1} is a crash course on basic \emph{statistical learning} concepts that provides motivations for the methods seen throughout the review. In chapter \ref{chapter2}, a classification model based on geometric separation of data, called \emph{support vector machines}, will be discussed. Support vector machines make use of a class of function known as \emph{kernels}, which act as a similarity measure between data points. Different kernels emphasize different properties of the underlying data, thus the task of choosing the right kernel for a given problem is of great interest. Chapter \ref{chapter3} is the main body of the review. Here, we give a broad overview of the literature of kernels defined on graphs, with an emphasis on kernels that may be useful in chemoinformatics. Our goal is to provide a list of popular graph kernels along with their properties, computational complexity, and test results so that the reader can become acquainted with many of the important terms, techniques, and papers in the literature so that they can determine if graph kernels can be helpful in their research.

\cleardoublepage 
\tableofcontents
\cleardoublepage 

\pagenumbering{arabic}
\setcounter{page}{1}



\chapter{Supervised Statistical Learning}
\label{chapter1}

In this first chapter, we briefly touch on what statistical learning is and what we are trying to accomplish with it. While mostly self-contained, it is not intended to be a comprehensive source on the subject's foundations. As such, we are only going to cover the most essential theoretical basics, as well as outline the general notation and terminology used throughout the review. For a more detailed and example driven treatment of this material, see the book \citep{james2021introduction} in which this chapter is based upon, or \citep{hastie2009elements} for a more advanced and mathematically rigorous approach. 

\smallskip

The general setting is the following: We receive a sample containing $n$ observations $y_i$ ($i\in\{1,\,\cdots , n\}$), each with a corresponding vector containing $p$ predictors (or \emph{inputs}) $\textbf{x}_i = (x_1,\,\cdots, x_p)^T$. Often, it will be useful to think of the observations and predictors as a set of ordered pairs $(\textbf{x}_i,\, y_i)_{i=1}^n$ inhabiting a higher dimensional space, with the $y_i$'s being the response to the inputs $\textbf{x}_i$. This notation will appear throughout the review. The $p$-dimensional space of predictors will be referred to as \emph{input space}. Now, given this sample, what can we say about other members in the population? In reality, it is not feasible to measure every single input of every member of an entire population. The basic idea of supervised statistical learning is to build a statistical model from our sample that allows us to predict observations based on inputs not found in our sample. We typically won't fit a model to \emph{all} of our sample. The subset in which we do is called the \emph{training data}, which is used to \emph{learn} how the predictors relate to the observations. The remaining members in the sample are apart of the \emph{testing data}. Inputs from this subset can be plugged into the model, allowing us to observe the predictive power of the model on new data. The term \emph{supervised} is in reference to the observations being present in our samples, which means we can measure the predictive power of our model by quantifying how close the predicted observations are to the true observations.  There is also what is known as unsupervised learning, where no response is observed from any predictor in the sample. In this situation, we have nothing to base predictions off of, so instead the analysis is based in understanding how particular observations relate to one another. Our focus for the duration of the review will be on supervised learning.

\smallskip

There are two types of observations we can encounter. Either the observations have quantitative values, meaning they are assigned a numerical quantity in which typical calculations make sense (eg.\ the real numbers $\mathbb{R}$), or they have qualitative values, meaning the value (or often label) assigned to them is meant to denote a classification, rather than a continuous quantity. Under this assumption, there is no meaningful way to combine the class labels as we do for quantitative variables.

\section{Regression Problems}
\label{Regression Problems}

Here, we begin our journey with the foundational elements of supervised statistical learning. The theory has been framed in terms of regression problems, although much of it applies to classification problems as well. The differences in the material will be highlighted in section \ref{Classification Problems} when we look at equivalent tools for classification problems.

\subsection{The Setup}

The question now is how do we build such a model described in the opening paragraph of this chapter? As per the naming convention, the predictors are thought to have a level of influence over their respective responses. We can model this relationship by writing
\begin{equation}
    y = f(\textbf{x}) + \varepsilon
\end{equation}

\noindent
where $f$ is an unknown function representing the relationship between $\textbf{x}$ and $y$ (or the information about $y$ provided by $\textbf{x}$), and $\varepsilon$ represents a random error term that is independent of $\textbf{x}$. This error term essentially accounts for unmeasured variables that, in theory, would be useful in predicting $y$. We assume that $\varepsilon$ has mean zero, as for larger and larger samples we expect the responses predicted by $f$ to over and underestimate $y$ equally. Of course in practice, $f$ is not known, so we must resort to estimating it. We represent this estimation by
\begin{equation}
    \hat{y} = \hat{f}(\textbf{x})
\end{equation}

\noindent
where $\hat{f}$ is the estimate of $f$ and $\hat{y}$ is its prediction for $y$. We can gain a theoretical understanding of the accuracy of our estimate by understanding the types of errors it introduces. The \emph{reducible error} is the error generated by $\hat{f}$ not being a perfect estimate of $f$. We call it reducible because if we came up with a better estimate, the error would decrease. The \emph{irreducible error} is the error generated by the variance of our random error term $\varepsilon$. It is irreducible because unlike $\hat{f}$, we have no control over it, and thus cannot make it smaller. Even if we could estimate $f$ perfectly so that $\hat{f} \equiv f$ and consequently $\hat{y}=f(\textbf{x})$, our prediction $\hat{y}$ would still have error built into it because $y$ also depends on the random variable $\varepsilon$, which is independent of $\textbf{x}$. Thus, it cannot be predicted by $f$. Symbolically, let us recall the expectation operator $E$. The expectation of a random variable $X$, written $E[X]$, is (informally) a measure of the average value of independent outcomes of $X$. Depending on context, this can involve a sum over all outcomes of $X$, each multiplied by their corresponding probability of occurring, or as an integral involving a probability density function of $X$, for example. These formulations imply that it is a linear operator that fixes constants. Returning to our discussion on the theoretical accuracy of $\hat{y}$ as a predictor of $y$, with a little algebra it can be shown that if $\hat{f}$ and $\textbf{x}$ are fixed, then
\begin{equation}
\label{reducible and irreducible error}
    E[(y-\hat{y})^2] = [f(\textbf{x})-\hat{f}(\textbf{x})]^2 + \text{Var}(\varepsilon)
\end{equation}

\noindent
where $\text{Var}(X)$ is the variance of a random variable $X$ defined as
\begin{equation}
    \text{Var}(X) \defeq E[(x-\mu)^2]
\end{equation}

\noindent
where $\mu = E[X]$ is the mean of $X$. Crucially for equation \ref{reducible and irreducible error} to hold, $\varepsilon$ has mean zero. The first term on the left-hand side of equation \ref{reducible and irreducible error} is the reducible error and the variance term is the irreducible error.


\subsection{General Modelling Approaches and Measuring Error}
\label{modelling and error}

The pressing question is: how do we produce a satisfactory estimate $\hat{f}$ for $f$? Furthermore, if we don't know what $f$ \emph{is}, how do we quantify the error introduced by our estimation? The unsatisfying answer for both is: it depends. While theory certainly is a factor when trying to estimate $f$ for a particular problem, often one must explore many different approaches to find one that works well for the problem and the sample available. There are two general classes of methods used to estimate $f$. The first are known as \emph{parametric methods}. Here, a common form for $f$ is assumed, such as a polynomial or a linear combination of elementary functions. For example, suppose that we think the relationship between predictors and responses is linear. We would then guess that $f$ has the form

\begin{align}
    f(\textbf{x}) &= \textbf{w}^T \textbf{x} + b \nonumber\\  &= w_p x_p + \cdots + w_1 x_1 + b.
\end{align}

Fitting the model with a method like linear least squares would yield the coefficients that result in the best fit for the training data, and thus produce an estimate for the true $f$. An advantage of this approach is that the problem of finding a function has been reduced to finding $p+1$ parameters. The downside is that the form is often too simple, and doesn't match the true $f$. One could trade in some of the simplicity by increasing the number of parameters, but the cost is that this can lead to \emph{overfitting} the training data. This phenomenon happens when the estimate is attempting to match the training data so closely that it picks up on patterns that only exist due to randomness in the training data (sometime called \emph{noise}). This issue often becomes apparent once one measures the prediction error of the testing data, as an overfit model tends to not produce good predictions.

\smallskip

The second option is to use a non-parametric method, which makes no assumption about the form of $f$ and simply attempts to balance matching the training data while remaining ``smooth", as in some differentiability condition imposed at training data points. With no restriction put on the shape of $f$, these methods can potentially match the true function $f$ more accurately than a parametric method. However, without imposing more restrictions, we require far more sample data with this method to be confident in an accurate prediction.

\smallskip

We won't go into any specific examples of the methods outlined above in this chapter, rather we will continue to examine the principles of statistical learning in generality. These methods are a prominent feature in many statistics textbooks. Chapter \ref{chapter2} is all about a hybrid method in the context of classification. Instead, we will begin to answer our second question of this section regarding the performance of our model on the test data. The function we will use for regression problems is the \emph{mean squared error} (MSE), which acts on a sequence of data points as follows:

\begin{equation}
\label{mse}
    (\textbf{x}_i,\,y_i)_{i=1}^m \mapsto \frac{1}{m}\sum_{i=1}^m [y_i-\hat{f}(\textbf{x}_i)]^2 .    
\end{equation}

Why do we chose this particular measurement, say over any other measurement such as a simple average of the absolute difference? There are a few reasons that may not be immediately apparent. For one, squaring the difference has the effect of minimizing the contribution of small errors, while amplifying the effect of large errors. The benefit of this is that it provides a simple way to increase the penalty associated with greater prediction errors, while being more lenient with small errors, since it is expected that our model will not be perfect in reality. Furthermore, squaring remains a relatively easy computation while being differentiable everywhere with respect to $\hat{y}=\hat{f}(\textbf{x})$. To be more precise, it is highly desirable to be able to differentiate an error function with respect to some parameter of the model $\hat{y}=\hat{f}(\textbf{x})$. For example, the linear regression model takes the form $\hat{y} = \textbf{w}^{T} \textbf{x} + b$. Thus differentiating with respect to $\textbf{w}$ can done to find the global minimum (as it is a convex optimization problem) of the MSE with respect to $\textbf{w}$, which is how $\textbf{w}$ is typically estimated in practice. If the data points in the input of the mapping \ref{mse} consist of training data, then we call this measurement the \emph{training MSE}. When they consist of the test data points, it is then called the \emph{test MSE}. Ultimately, we do not care as much about the training MSE. We want to make predictions, so our goal is to minimize the test MSE. Moreover, it is not generally true that minimizing training MSE will result in a minimum test MSE. A fundamental truth in statistical learning is that as model flexibility increases (i.e.\ how much the model is influenced by the given data), we tend to observe a monotonic decrease in the training MSE, but the test MSE traces out a characteristic \emph{U}-shape. Starting small, as flexibility begins to increase, the test MSE decreases towards a minimum before increasing again. With our new terminology, we can restate the phenomenon of overfitting data as obtaining a small training MSE, but a large test MSE, which tends to happen for a model that is very flexible. Why does the \emph{U}-shape appear in the test MSE? It comes from competing properties known as \emph{bias} and \emph{variance}. Let $\hat{\theta}$ be an estimator of a fixed parameter $\theta$. If we define the MSE of $\hat{\theta}$ as

\begin{equation}
    \text{MSE}(\hat{\theta}) \defeq E[(\hat{\theta}-\theta)^2]
\end{equation}

\noindent
and define the bias of the estimator $\hat{\theta}$ as 

\begin{equation}
    \text{Bias}(\hat{\theta}) \defeq E[\hat{\theta}] - \theta
\end{equation}

\noindent
then it can be shown that 

\begin{equation}
    \text{MSE}(\hat{\theta}) = \text{Bias}(\hat{\theta})^2 + \text{Var}(\hat{\theta})
\end{equation}

\noindent
which shows that a decrease in bias is offset by an increase is variance, and vice-versa. In our situation where we are measuring the expected test MSE of a learning method for a fixed observation pair $(\textbf{x}_0,\,y_0)$, this Bias-Variance trade-off is found in the equation

\begin{align}
    \text{Expected Test MSE} &\defeq E[(y_0 - \hat{f}(\textbf{x}_0))^2] \nonumber \\ &= \text{Bias}(\hat{f}(\textbf{x}_0))^2 + \text{Var}(\hat{f}(\textbf{x}_0)) + \text{Var}(\varepsilon)
\end{align}

\noindent
which can be derived in a similar way as equation \ref{reducible and irreducible error}. The squared bias and the variance of the estimate in this equation are always non-negative, so the lower bound on the expected test MSE is the variance of the random noise term $\text{Var}(\varepsilon)$. Intuitively, the expected test MSE at $\textbf{x}_0$ is the value that the average test MSE at $\textbf{x}_0$ would approach if we repeatedly estimated $f$ on more and more training sets. The bias of a statistical learning method represents the error introduced by approximating a (possibly complex) real-life problem by a (possibly simplistic) model. An example where high bias can occur is if we use linear regression to model a relationship between variables that is truly not linear. Think of this in terms of the colloquial use of the term bias. The model is essentially ignoring certain information and only looking to validate the assumption that a linear relationship holds. Variance on the other hand is a measure of how much our $\hat{f}$ would change if we followed the same procedure for finding it, but instead used a different training set for learning.

\smallskip

We can now explain the general behaviour of the test MSE. The relative rate of change of bias and variance dictates whether test MSE will rise or fall. Starting with low flexibility, typically what we will see as the flexibility of a model increases is an initial rapid decrease in bias in comparison to the increase in variance, corresponding to a decrease in expected test MSE. At some point however, the increasing flexibility of the model will yield diminishing returns on lower bias, but will cause variance to rise dramatically as the model begins overfitting the data, corresponding to an increase in expected test MSE. This is what gives the plot a characteristic U-shape. In reality, the true $f$ is not known, so we have no way to compute the expected test MSE, the bias, or the variance. It is still important to be aware of them as they form the theoretical underpinning of certain model behaviour. We do however have methods to estimate the test MSE, which we will see shortly.


\subsection{Resampling}
\label{Resampling}

Continuing with the theme of exploring model error, we now turn our attention to a fundamental problem in statistics: sample size. The situation of not having ``enough" members in a sample in very prominent, and attempts to rectify some of the issues it presents is of great interest. One occurrence of this issue is known as the \emph{curse of dimensionality}. While this ominous term very generally refers to the differences between the properties of high and low-dimensional data, a particular instance as it relates to sampling is where increasing the number of inputs (dimension) causes an exponential increase in the ``size" or ``volume" of the input space. What ends up happening is that a correspondingly exponential increase in sample size is needed to accommodate this increase in volume, as we would like for our model to have seen many different possible combinations of inputs (otherwise we risk having high variance).

\smallskip

Consider as before a partition of the samples into a training and a testing set. The way that the partition is formed is random in an effort to mitigate any implicit biases such as the way the data was collected. But what if our random choices ends up selecting a subset that is not very representative of the sample as a whole? The techniques we are going to describe are known as \emph{resampling methods}. These involve drawing many different partitions of our sample, and fitting the model each time on the new training sets in order to gain more accurate information about the performance of the model. Essentially, this is a sort of smoothing process that seeks to lessen the obscuring effect that a random partition can have on our measurement of the test error. We are going to look at two very common methods for resampling in statistical learning. The first method is called \emph{cross-validation} (CV). It works by holding out on a subset of the observations during each iteration to use as the testing set, while the remaining observations are used for training. There are a few approaches to implementing cross-validation we will discuss, each of which is defined by the size of the holdout set.

The first and simplest approach is the \emph{validation set} approach. During each resampling, we simply take half of the observations for the training set, and the other half for the validation (testing) set. As is typical, the selections for each set is random after deciding on the proportions. Once this is completed, the model can be fit to the training data and the \emph{validation error rate} can be computed as an estimate for the test error rate. This is typically done using the MSE function. As it stands, the validation set approach performs poorly compared to its modified counterparts, as the validation error rate can be highly variable. In practice, statistical models tend to perform worse when trained with fewer observations. What results is a tendency to overestimate the test error rate of the data.

On the other side of the spectrum, we have \emph{leave-one-out cross-validation} (LOOCV). During each iteration of the resampling, we again randomly partition the sample. This time however, only a single observation-predictor pair is selected for the validation set, say $(\textbf{x}_1,\,y_1)$, which leaves the rest for training. We fit the model on the $n-1$ observations in the training set and then compute the test MSE for this iteration, which is given by $\text{MSE}_1 = (y_1 - \hat{y}_1)^2$. This process is repeated, each time selecting a different observation for the validation set and obtaining another estimate for the test error. After exhausting all $n$ unique validation sets, we take the average of each of the $n$ MSE terms as the LOOCV estimate for the test error:

\begin{equation}
    \text{CV}_{n} \defeq \frac{1}{n} \sum_{i=1}^n \text{MSE}_i .
\end{equation}

As each $\text{MSE}_i$ term only makes its measurement on a single observation and the model is trained on nearly the entire sample set, it follows that each MSE calculation is an approximately unbiased estimate for the true test error. Mathematically, this can be shown by first assuming our model is correct (a big assumption typically made for theoretical calculations) and fixing an observation-predictor pair $(\textbf{x}_0,y_0)$. Then we have that $E[y_0] = \hat{y}_0$, and using the fact that $E[\varepsilon] = 0$, it can be shown that
\begin{align}
    E[\text{MSE}_0] &= E[(f(\textbf{x}_0) + \varepsilon - \hat{f}(\textbf{x}_0))^2] \nonumber\\ &= E[\varepsilon^2] \nonumber\\ &= \text{Var}(\varepsilon)
\end{align}

\noindent
and thus
\begin{equation}
    \text{Bias}(\text{MSE}_0) = E[\text{MSE}_0] - (\text{True Test Error}) = 0 .
\end{equation}

However, there is a large variance associated with this estimation. It can be reduced slightly when we compute the overall estimate $\text{CV}_n$, but this comes at the cost of added computational time for fitting a model $n$ times. The result of the decreased bias is that the LOOCV estimate doesn't overestimate the test error as much as the validation set approach. Moreover, as the validation sets are picked systematically, the LOOCV method has no randomness in its partitioning, versus the validation set approach which yields different results each time it is applied. The LOOCV method is very general, and can be paired with nearly any kind of predictive model.

As mentioned, one drawback to LOOCV is that fitting a model $n$ times can be computationally expensive \textemdash especially when $n$ is large or the model is complex. The next method attempts to deal with this drawback, along with finding a more desirable balance of bias and variance, while still attempting to capture the essence of LOOCV. 

\begin{remark}
In the case that one uses linear or polynomial regression, a remarkable formula found in \citep[ch.\ 5]{james2021introduction} computes the LOOCV estimate for the test error $\text{CV}_n$ with the same computational cost as one model fit, as opposed to $n$ model fits.
\end{remark}

The final approach we will look at, called \emph{k-fold cross-validation}, is one generalization of LOOCV that performance-wise inhabits the space between the validation set approach and LOOCV. For a natural number $k$, we randomly divide the set of $n$ observations into $k$ groups (called \emph{folds)} of roughly equal sizes. Choosing one fold at a time for the validation set, we fit our model to the $k-1$ remaining folds, and then compute the MSE using the validation fold. This yields $k$ estimates for the test error, and we use the average of these quantities as the $k$-fold CV estimate:

\begin{equation}
    \text{CV}_k \defeq \frac{1}{k} \sum_{i=1}^k \text{MSE}_i .
\end{equation}

Notice that LOOCV is a special case of this approach, corresponding to $n$-fold CV. In practice, it is common to see 5 or 10-fold implementation, 5 for smaller data sets (say $n<100$) and 10 for large data sets ($n>1000$). These values tend to strike a good balance between training with lots of observations, while leaving enough for testing to get an adequate estimation of performance. As suggested previously, the $k$-fold method performs somewhere between the other, more extreme methods. While LOOCV tends to have lower bias due to the number of training observations being maximal, the viewpoint of modern theory is that in practice, it can be better to have a biased estimate with smaller variance. Recalling the bias-variance trade-off, LOOCV will indeed have higher variance that $k$-fold CV. This can be further explained due in part to the high positive correlation between the outputs of LOOCV. Notice that the training sets in LOOCV are nearly identical from iteration to iteration, and hence knowing the performance of a model trained on one gives us a good idea of the performance of the same model trained on the others. In contrast, outputs in $k$-fold CV are less positively correlated due to having more varied training sets, and hence a comparatively smaller variance. This can be expressed mathematically using the identity for the variance of the sum of random variables. Let $F_i$ and $F_j$ represent two different folds with randomly chosen members. Then

\begin{equation}
    \text{Var}(F_i + F_j) = \text{Var}(F_i) + \text{Var}(F_j) + 2\text{Cov}(F_i,F_j).
\end{equation}

We see that the variance becomes smaller when the covariance term is closer to 0. As a final note, it is clear to see that $k$-fold CV requires far less computational time to execute when compared to LOOCV. As important as test error is to assessing models, its actual value may not always be of interest. If we want to identify a method or level of flexibility that results in lowest test error, it really doesn't matter what this value is. Rather, we want to know the \emph{location} of the minimum test error in terms of level of flexibility, either across multiple curves (in the flexibility-MSE plane) representing different methods, or along the same curve representing one method.

\smallskip

The second, very general method for resampling we will briefly mention is the \emph{bootstrap} method. It is a widely employed tool for quantifying the accuracy of estimators and statistical methods. It finds many use cases, especially when these measurements are quite difficult to compute or statistical software doesn't compute them automatically. The outline for how it is performed is as follows: Choose $n$ observations from the sample set of size $n$ with replacement, meaning the same observation may be chosen multiple times, or not at all. This is called the bootstrap data set. With this new data set, we make the measurement or calculate the statistic that we ultimately want to estimate. This process is repeated many times, and the mean of theses values is used as the bootstrap estimate for its true value. The effect of resampling using the bootstrap method is that it simulates the process of gathering new samples from the population.


\section{Classification Problems}
\label{Classification Problems}

Next, we will touch on some basic supervised statistical learning practices as it pertains to classification problems. Recall that the distinction with classification problems is the observed responses are placed into categories rather than given a continuous numerical value. Most of the concepts from section \ref{Regression Problems} on regression problems remain largely the same, with differences having a natural ``discrete" analogue. We will mainly touch on how some of the tools change in this setting, as the core statistical principles are essentially invariant to the domain of the observations.

\smallskip

Suppose that we wish to estimate $f$ on the basis of a subset of our observations-predictor pairs as we did in the previous section, except now the observations $y_i$ are \emph{qualitative}. We call the estimator $\hat{f}$ for $f$ a \emph{classifier}, as it takes a vector containing $p$ features as input, and outputs a class label prediction (typically represented with an integer). The function we will use to quantify the empirical error introduced by $\hat{f}$ is defined by the following map:

\begin{equation}
\label{classification error}
    (\textbf{x}_i,\,y_i)_{i=1}^m \mapsto \frac{1}{m}\sum_{i=1}^m \textbf{I}(y_i \neq \hat{y}_i).
\end{equation}

The function $\textbf{I}(y_i \neq \hat{y}_i)$ is called the \emph{$0\text{-}1$ loss function}, where \textbf{I} denotes the indicator variable that outputs 1 if the input is true and 0 otherwise. Thus, the function defined by the mapping in equation \ref{classification error} can be used to compute the ratio of incorrect classifications made by $\hat{f}$. The \emph{training error rate} replaces the training MSE from section \ref{Regression Problems}, and is computed using the mapping in equation \ref{classification error} with the training set as input. Similarly, the \emph{test error rate} replaces the test MSE and is found by computing the mapping in equation \ref{classification error} on the test set. Once again, we are most interested in reducing the test error rate rather than the training error rate, as our goal is to make predictions for new data.

\smallskip

How can we use the training data to find a classifier? In theory, what is known as the \emph{Bayes} classifier would be optimal in the sense that on average, it is the classifier that minimizes the test error rate. It works simply by assigning each observation to the class that it most likely belongs to given its inputs. That is, a predictor $\textbf{x}_0$ is assigned to the class $j$ that maximizes the conditional probability $P(y=j\mid \textbf{x}=\textbf{x}_0)$. If we have $r$ possible classifications, we could define the Bayes classifier by the mapping

\begin{equation}
    \textbf{x}_0 \mapsto \argmax_{j\in\{1,\cdots,r\}}P(y=j\mid \textbf{x}=\textbf{x}_0) .
\end{equation}

A derivation of the Bayes classifier can be found in \citep{hastie2009elements}. When there are only two possible classifications, the set of points in the input space for which $P(y=j \mid \textbf{x}=\textbf{x}_0) = \frac{1}{2}$ is called the \emph{Bayes decision boundary}. This boundary splits the input space between the two classes, although the set of points representing a class need not be connected. For the points satisfying $P(y=j \mid \textbf{x}=\textbf{x}_0) > \frac{1}{2}$, the classifier will put them into class \emph{A}, and for points satisfying $P(y=j \mid \textbf{x}=\textbf{x}_0) < \frac{1}{2}$, it will put them into class \emph{B}. One can visualize this by imagining a plane where each axis represents a particular input (i.e.\ the input space is two dimensional), then splitting the plane with a curve (which will be the decision boundary). We can imagine the members of class \emph{A} inhabiting one side of the curve, and the members of class \emph{B} inhabiting the other side. Do note that the Bayes classifier is the best at minimizing test error on average, and is not perfect. With real data, some of the observations may lie on the ``wrong side" of the decision boundary. Moreover, in general we won't always be able to cleanly divide the observations into distinct groups. The Bayes error rate at a point $\textbf{x}_0$ is given by $1-\max_j P(y=j\mid \textbf{x}=\textbf{x}_0)$, and the overall Bayes error rate is $1-E[\max_j P(y=j\mid \textbf{x})]$, where the expectation is taken with respect to $\textbf{x}$. The Bayes error rate is a form of irreducible error, just as $\textup{Var}(\varepsilon)$ was in section \ref{Regression Problems}. As mentioned previously, the Bayes classifier remains an unobtainable theoretical best. This is due to the conditional probability found in its definition, which cannot be computed as we do not know the conditional distribution of $y$ given $\textbf{x}$. Next, we give an example of a simple method that estimates this conditional probability.

\smallskip

Given a test observation $\textbf{x}_0$, it is quite natural to base the prediction for its classification on nearby training observations. This is the core idea behind the \emph{K}-Nearest Neighbours (KNN) classifier. First, we decide on how many neighbours $K$ can influence classification. Next, it finds the closest \emph{K} points to $\textbf{x}_0$. This requires a notion of distance, which in this context we simply take as the regular Euclidean distance. We label this set of neighbours with the notation $N_K(\textbf{x}_0)$. It then estimates the probability that $\textbf{x}_0$ is in the \emph{j}th class using the following as an approximation:

\begin{equation}
    P(y=j\mid \textbf{x}=\textbf{x}_0) \approx \frac{1}{K} \sum_{\textbf{x}_i\in N_K(\textbf{x}_0)} \textbf{I}(y_i = j).
\end{equation}

Now, Bayes rule of classification can be applied, again picking the class for which the estimated probability is highest. KNN classification tends to work quite well in practice. One can adjust the flexibility of the KNN classifier by simply choosing different values for $K$. Remember that we are most interested in minimizing the test error and not so much the training error. Thus to achieve the best results, we can apply the method numerous times while varying $K$ to find the one in which test MSE is minimized.

\smallskip

Finally, we may want to employ resampling methods to gather more information from our samples. Cross validation, as explored in section \ref{Regression Problems}, works in much the same way as we saw before. Only this time, our estimate for the test error uses the training error rate found in equation \ref{classification error} instead of MSE.


\chapter{Support Vector Machines}
\label{chapter2}

Given some characteristics of an object, what from those characteristics differentiates it from other, similar objects?  For example, how does your email provider decide whether a given message belongs in your inbox, or in the spam folder? How does a bank determine which transactions are fraudulent? Based on patient symptoms and data collected from samples, can we say with any level of confidence if a tumor is cancerous or benign? Classifications problems are ubiquitous, and having a general, easy to implement, and efficient way of solving these problems is crucial for many aspects of modern-day life. At the end of chapter \ref{chapter1}, we briefly touched on an approach to classification known as $K$-nearest neighbours that classified observations based on the prominent class amongst similar observations. This was in an attempt to approximate the optimal Bayes classifier, which itself is pulling from the true conditional distribution of the predictor and response variables. We will now explore an alternative approach to classification using what are known as \emph{support vector machines}. In essence, support vector machines make one assumption of the given data, which is that there is level of separability of the classes. This differentiates them from the $K$-nearest neighbours classifier as no assumption about a distribution is made. Rather, they seek to partition classes from a geometric point of view. Picture it as drawing a line or curve through the data, keeping an equal distance from both classes, instead of focusing on particular points and factoring in the local behaviour of other points. We mainly concern ourselves with the case of binary classification problems, which is the setting in which support vector machine were created and often employed. There have been attempts to extend the theory to a greater number of classes, which will be briefly mentioned at the end of the chapter.

\smallskip

The chapter will proceed roughly as follows. First, we will start with a light introduction to the concept of classifying observations based on ``separation" by assuming that our observations can be perfectly separated into two distinct classes by a linear function called a \emph{hyperplane} (section \ref{Separable Observations and the Maximal Margin Classifier section}). In practice, this method by itself is not very useful since classes tend to overlap, so we will then move to the \emph{support vector classifier} that attempts to deal with observations that seem to cross what we would perceive to be the decision boundary (section \ref{Non-Separable Observations and the Support Vector Classifier section}). This support vector classifier is but a special case of the general \emph{support vector machine}, or \emph{SVM} (section \ref{Nonlinear Classifiers section}). It is not always true that a linear decision boundary accurately describes where one class should end and another should begin at every point in space. In general, SVMs replace such a ``dividing line" with one or more curves that can bend around data points in the space. It will be here that we discuss a very special class of functions known as ``kernels", which measure how similar two observations are in some very specific way. We will assume that our input space is the real vector space $\mathbb{R}^p$. This presentations is adapted from \citet[ch.\ 9]{james2021introduction}, \citet[ch.\ 11]{theodoridis2015machine}, and \citet{moguerza2006support}.


\section{Hyperplane Classification}
\label{Hyperplane}

For a $p$-dimensional space (eg. the real vector space $\mathbb{R}^p$), a \emph{hyperplane} is a $p-1$ dimensional affine subspace. The term ``affine" means that the space need not pass through the origin, as a typical vector subspace would, but must still retains its closure properties. For example, a hyperplane $H$ in $V$ could be written as $H = \textbf{v}_0 + W$ for some vector $\textbf{v}_0$ in $V$, and subspace $W$ of $V$. In the regular $xy$-plane, a hyperplane would be any straight line dividing the plane. Similarly in $\mathbb{R}^3$, a hyperplane would look like a two dimensional flat sheet dividing the space, or in other words a ``copy" of $\mathbb{R}^2$ inside $\mathbb{R}^3$. More precisely, we can parameterize a hyperplane so that it can be written as the set of points $\textbf{x} = (x_1,\cdots,x_p)^T$ in the space satisfying
\begin{equation}
    \textbf{w}^T \textbf{x} + b = w_p x_p + \cdots + w_1 x_1 + b = 0
\end{equation}

\noindent
for some fixed vector $\textbf{w} = (w_1,\,\cdots,w_p)^T$ and constant $b$. The intuitive picture of the dividing nature of the hyperplane can be formulated mathematically by saying the set of points satisfying $\textbf{w}^T \textbf{x} + b > 0$ are on one side of hyperplane, while the set of points satisfying $\textbf{w}^T \textbf{x} + b < 0$ are on the other side. When such a hyperplane exists, a test data point $\textbf{x}_i$ can be classified based on the sign of $\textbf{w}^T \textbf{x}_i + b$. Thus, we represent the two classes with $y\in \{1,-1 \}$. The quantity $\labs \textbf{w}^T \textbf{x}_i + b\rabs$ also carries with it useful information. The larger its value, the further away from the hyperplane the observation will lie, and hence the more confident we are in the correctness of the classification.

\smallskip

Next we establish some terminology and common assumptions regarding hyperplane classification. Recall that the perpendicular distance from a point $\textbf{x}_0$ to a hyperplane is given by the quantity
\begin{equation}
    \frac{| \textbf{w}^{\,T} \textbf{x}_0 + b |}{\| \textbf{w} \|}.
\end{equation}

As the hyperplane is invariant to scaling, if $\textbf{x}_0$ is the closest training data point to the hyperplane, we can arrange for $| \textbf{w}^T \textbf{x}_0 + b | = 1$ so that the perpendicular distance is $1/\|\textbf{w}\|$. This ``normalizing" of the hyperplane will be assumed henceforth. The goal is to build an \emph{optimal margin classifier}, meaning the classifier doesn't favour one class over the other. The optimal hyperplane will have at least two points that are ``closest" to it, that is, two points with perpendicular distance $1/\|\textbf{w}\|$ to the hyperplane. These points lie on either side of the hyperplane and represent the closest points in each training data class to the hyperplane. Thus, the distance between the classes, called the \emph{margin}, is $2/\|\textbf{w}\|$. The two hyperplanes defined by $\textbf{w}^T \textbf{x} + b = \pm 1$ are together called the \emph{boundary of the margin}. The goal can be stated more concretely as maximizing the margin, or equivalently, minimizing $\|\textbf{w}\|$. 


\subsection{Separable Observations and the Maximal Margin Classifier}
\label{Separable Observations and the Maximal Margin Classifier section}

With the general idea of hyperplane classification in mind, we now begin with the simplest approach. This occurs when the data is \emph{linearly separable}, meaning there exists a hyperplane such that one side of it contains only one class and the other side contains only the other. Our goal is to construct an optimization problem that can find a vector $\textbf{w}$ and constant $b$ parameterizing a hyperplane that sits perfectly between the two classes, meaning that no training point is misclassified and that the plane is equidistant from the nearest points in each class. This process will serve as a template for each successive, more general classification procedure. For the reader looking for a more thorough exploration of the theory, a standard source for (convex) optimization theory is \citet{boyd2004convex}. Outlines of the solutions to the optimization presented here can be found in \citet[ch.\ 11]{theodoridis2015machine}.

\smallskip

For a fixed training set $(\textbf{x}_i,y_i)_{i=1}^m$, we want to train a model $\hat{f}$ (also called a decision rule or discriminant function in the context of classification problems) that minimizes the empirical loss from misclassification
\begin{equation}
\label{empirical loss}
    \hat{f} \mapsto \sum_{i=1}^m L(y_i, \hat{f}(\textbf{x}_i))
\end{equation}

\noindent
where $L$ is a \emph{loss function}. We saw an example of a loss function in section \ref{Classification Problems}, which was the 0-1 loss function. A different loss function will be used to construct SVMs for a few reasons. While the 0-1 loss function is very simple and quite natural, the presence of jump discontinuities make it incompatible with many optimization methods that require levels of differentiability. It is possible to find a work around if a loss function has points where it is only continuous by using a smooth function to approximate the loss function, but this isn't viable for the 0-1 loss function since it is discrete. Instead, the \emph{hinge loss function} is used:
\begin{equation}
    L_H(y_i, \hat{f}(\textbf{x}_i)) \defeq \max(0,\, 1 - y_i \hat{f}(\textbf{x}_i)).
\end{equation}

Since we know that the sign of a normalized hyperplane will be used for $\hat{f}$, it is immediately apparent how $L_H$ will penalize classifications. A penalty is incurred whenever $y_i \hat{f}(\textbf{x}_i) \leq 1$, corresponding to a violation of the margin. Otherwise, the point has been correctly classified, so no penalty is added. What this loss function does is give us a \emph{sparse solution}, since the only points that are going to influence the hyperplane are the points lying on the boundary of the margin, which generally consists of a very small subset of the training data. This is desirable as it makes the classifier resilient to outliers and it gives them an advantage when dealing with large amounts of data. Thus, the problem of designing an optimal classifier with the desired properties can be recast as finding the minimum of the following cost function:

\begin{equation}
    (\textbf{w}, b) \mapsto \frac{1}{2} \| \textbf{w} \|^2 + C\sum_{i=1}^n L_H (y_i,\, \textbf{w}^{T} \textbf{x}_i + b).
\end{equation}

Let us dissect the meaning of this cost function. First, as mentioned at the beginning of section \ref{Hyperplane}, we want to maximize the margin subject to some constraints. This is equivalent to minimizing $\| \textbf{w} \|^2$, as squaring the norm doesn't change the location of the minimum. The squaring is done to simplify the computation. The factor of $\frac{1}{2}$ is to cancel the factor of 2 that results from differentiation. This scaling has no effect on the second term, as it itself is scaled by a user-controlled hyperparameter $C>0$ that controls the weight of the accumulated losses. In the case of linearly separable data, minimizing this cost function is equivalent to the following optimization problem

\begin{equation}
\begin{aligned}
\label{linearly separable optimization problem}
\text{minimize} \quad\quad & \frac{1}{2}\| \textbf{w} \|^2 \\
\text{subject to} \quad\quad & y_i(\textbf{w}^T \textbf{x}_i + b) \geq 1,\, i \in \{1,2,\, \cdots, n\}.\\
\end{aligned}
\end{equation}

The solution is given by the finite linear combination

\begin{equation}
    \textbf{w} = \sum_{i\in \mathcal{S}} \lambda_i y_i \textbf{x}_i
\end{equation}

\noindent
where $\mathcal{S}$ is the set of indices of the nonzero Lagrange multipliers $\lambda_i$ obtained from the dual representation of the problem. Each Lagrange multiplier is associated to a constraint $y_i(\textbf{w}^T \textbf{x}_i + b) \geq 1$, and remarkably, the nonzero Lagrange multipliers correspond to precisely those constraints for which $y_i(\textbf{w}^T \textbf{x}_i + b) = 1$, which as described earlier are the closest points to the hyperplane lying on the boundary of the margin. These training points are called \emph{support vectors}, as they completely determine the hyperplane. In effect, the solution is able to ignore all but the most essential observations. Now, \emph{b} can be found by taking one of the constraints for which $\lambda_i \neq 0$ ($i \in \mathcal{S}$) and solving
\[
y_i (\textbf{w}^T \textbf{x}_i + b) = 1
\]

\noindent
for $b$. In practice, different constraints may have slightly different values of $b$ due to rounding errors, so for the sake of numerical stability, the average over the set $\mathcal{S}$ is taken. Therefore, our classifier $\hat{f}$ predicts the class $\hat{y} \in \{1,-1\}$ for a test observation $\textbf{x}$ using the formula

\begin{equation}
\label{support vector classifier}
    \hat{f}(\textbf{x}) = \sgn \left(\sum_{i\in \mathcal{S}} \lambda_i y_i  \left(\textbf{x}^T \textbf{x}_i\right) + b\right)
\end{equation}

\noindent
where $\sgn$ is the sign function that outputs 1 if the input is positive, and -1 if the input is negative. The function in equation \ref{support vector classifier} is what is known as the \emph{maximal margin classifier}.


\subsection{Non-Separable Observations and the Support Vector Classifier}
\label{Non-Separable Observations and the Support Vector Classifier section}

Next, we describe the procedure for dealing with two classes that are not linearly separable. We can imagine a scenario where their are two concentrations of observation in our data, but the border between one and the other is not entirely clear due to noise. Some observation appear to be on the ``wrong side" of the plot. For now, we are still only considering the case where a linear classifier is most appropriate. To codify this, we need an optimization problem with a built-in level of tolerance for how many training observations can be misclassified, and by how much. This is accomplished through the following optimization problem:
\begin{equation}
\begin{aligned}
\text{minimize} \quad\quad & \frac{1}{2}\| \textbf{w} \|^2 + C \sum_{i=1}^n \varepsilon_i \\
\text{subject to} \quad\quad & y_i(\textbf{w}^T \textbf{x}_i + b) \geq 1-\varepsilon_i,\\
&\varepsilon_i \geq 0 ,\, i \in \{1,2,\, \cdots, n\}.
\end{aligned}
\end{equation}

This problem will maximize the margin around the decision boundary, while ensuring a minimum number of misclassifications. In this case, we say that there is a soft-margin, meaning that observations may cross it, but at a cost. The user-defined variable $C\geq 0$ is a hyperparameter that acts like a budget. It give an upper bound on how many violations we are willing to accept, and how severely. When $C=0$, the problem is reduced to equation \ref{linearly separable optimization problem} corresponding to the separable case. When $C>0$, then at most $\lfloor C\rfloor$ observations are allowed to cross the hyperplane, and consequently be misclassified. An increase in $C$ will widen the margin, while a decrease in $C$ will shrink the margin. To tie this idea back to our discussions in chapter \ref{chapter1}, $C$ is controlling the bias-variance trade-off of the method. A larger $C$ allows for more violations of the margin, which generally results in a high bias but a low variance. Conversely, a smaller $C$ allows for fewer violations, so the classifier will be highly fit to the training data, which suggests a low bias but a high variance.

The solution to this optimization problem takes the same form as in the separable case. The only difference is which constraints have a nonzero Lagrange multiplier. Let us briefly describe the three types of observations and the values of their associated Lagrange multipliers. A predictor $\textbf{x}_i$ whose constraint satisfies $y_i(\textbf{w}^T \textbf{x}_i + b) \geq 1$ has been classified correctly and lies on the boundary or outside of the margin. These points incur no penalty, so $\varepsilon_i = 0$ and $\lambda_i = 0$. Now, if $0 < y_i(\textbf{w}^T \textbf{x}_i + b) < 1$, then $\textbf{x}_i$ has been classified correctly, but lies inside the margin. These points are said to violate the margin, and incur a penalty $0<\varepsilon_i<1$. Their associated Lagrange multiplier in this case are nonzero. Finally, if $y_i(\textbf{w}^T \textbf{x}_i + b) \leq 0$, then $\textbf{X}_i$ has been incorrectly classified. These points incur a penalty $\varepsilon_i>1$, and again have a nonzero Lagrange multiplier. In this context, not only are the points on the boundary of the margin support vectors, but so are all points within the margin and outside of the margin on the wrong side of the classifier. These classifiers are commonly called \emph{support vector classifiers}, although the terminology in the literature may simply call them support vector machines for simplicity. We reserve this title for the more general classifier in the next section.


\section{Nonlinear Classification}
\label{Nonlinear Classifiers section}

In the previous section, we saw how a classifier can be constructed using a linear combination of dot products with support vectors. We typically think of the dot product as a similarity measure, taking into account the angle between two points through the identity $\textbf{x}^{\,T} \textbf{y} = \| \textbf{x} \| \| \textbf{y} \| \cos{\theta}$, with $\theta$ being the angle between the points. A dot product is a particular example of an \emph{inner product}, which is a function used to quantify this same type of similarity in general vector spaces. This fact is used to construct support vector machines that solve the task of classifying data that is not only nonseparable, but also where a nonlinear decision boundary will lead to a lower test error.


\subsection{Reproducing Kernel Hilbert Spaces}

Hilbert space is a very important generalization of Euclidean space found throughout Mathematics, Physics, and Statistics. It is a special type of vector space endowed with many of the familiar notions of distance, angle, projection, and ``continuity". More formally, a Hilbert space $\mathbb{H}$ is a \emph{complete inner product space}. This means that it is a complete vector space (possibly infinite dimensional) over a field of scalars (typically the real or complex numbers) equipped with an inner product. An inner product is a scalar-valued function that gives a way to quantify similarities between vectors. Now, inner products induce a norm (given by the square root of the inner product), which is interpreted as a measure of length of vectors, and can be used to define distance between vectors. The final piece of the definition is \emph{completeness}, meaning that any sequence of points in the space that get arbitrarily close together with respect to the norm must also get arbitrarily close (or converge) to an element in the space. Sequences with this property are known as \emph{Cauchy sequence}. This can intuitively be thought of as the space having ``no holes". A familiar example of a complete space is the real numbers with the Euclidean metric. This space can be viewed as the completion of the space of rational numbers, which is famously incomplete, not containing fundamental constants such as $\sqrt{2},\,\pi,\, \textup{and } e$. Here we are only interested in Hilbert spaces consisting of real-valued functions defined on a subset of $\mathbb{R}^n$, plus an extra bit of structure described in the following definition.

\begin{definition*}
A Hilbert space of real-valued functions $\mathbb{H}$ on a set $U\subset\mathbb{R}^n$ is called a Reproducing Kernel Hilbert Space (RKHS) if there is a real-valued function $k: U \times U \to \mathbb{R}$ with the following properties:

\begin{enumerate}[i]
    \item\!\!. For every point $x$ in $U$, the function $k_x(\cdot) \defeq k(\cdot,x)$ is in $\mathbb{H}$.
    \item\!\!. For every function $f$ in $\mathbb{H}$ and point $x$ in $U$, we have that $f(x) = \langle f,\, k_x(\cdot) \rangle$.
\end{enumerate}

Property ii. is called the reproducing property of k, and k is called the reproducing kernel of $\mathbb{H}$.
\end{definition*}

Notice that the function $k$ and the inner product can be used to evaluate any function in the space at any point. A consequence of the reproducing property of $k$ is that for any points $x,\,y \in U$,
\begin{equation}
\label{kernel property}
  k(x,y) = k(y,x) = \langle k_x(\cdot), k_y(\cdot) \rangle.  
\end{equation}

The mapping $x \mapsto k_x(\cdot)$ is called the \emph{feature map}, and in combination with the above inner product, will ultimately allow us to access the rich structure of the RKHS \emph{implicitly}. The procedure for this is as follows: Starting with a set containing training data $(\textbf{x}_i,\,y_i)_{i=1}^m$, the feature map is employed implicitly through equation \ref{kernel property} by evaluating the kernel function $k(\textbf{x}_i,\textbf{x}_j)$ on the left-hand side. This has the effect of measuring the similarity of the two functions in the RKHS associated to the training points, which can be leveraged in classification problems as the training points might not be linearly separable in the input space, but the associated functions might be in the RKHS. We know that this approach is potentially viable due to Cover's theorem, which proves that data becomes arbitrarily separable as the number of features becomes arbitrarily large (\citet{cover1965geometrical}, \citet{theodoridis2015machine}). In the context of support vector machines, this RKHS is called the \emph{feature space}. The feature space typically has a much higher dimension than the input space (sometimes an infinite number of dimensions), and the feature map arrives at these extra dimensions by effectively building new predictors out of the ones in the input space. Next, we will highlight some relevant technical details of RKHS's.

\begin{proposition*}
Let $\mathbb{H}$ be a RKHS on a set U with kernel k. Then

\begin{equation}
\mathbb{H} = \overline{\spn\{k_x(\cdot) : x\in U\}} .
\end{equation}

\end{proposition*}

What this result is saying is that any function in the RKHS can be generated from the kernel function, either by a finite linear combination of $k_{x_i}$'s, or by an infinite series. The idea behind why this is true is that the only function orthogonal to all elements in the span of the $k_x$'s is the zero function, due to the reproducing property. Then, one can use the well-known result that any Hilbert space can be decomposed into the direct sum of any closed subspace and its orthogonal complement \citep[see][Theorem 12.4]{rudin1991functional}.

\smallskip

It is important to note that RKHS's have a few different characterizations, and one may encounter them from sources using an alternative definition as a starting point. \citep{scholkopf2002learning} contains complete presentation of these characterizations. We will briefly cover some of these in the remaining paragraphs of this section. An equivalent (and more intuitive) formulation of an RKHS is that every evaluation functional defined on the entire space is continuous \citep{aronszajn1950theory}. In other words, two functions that are close with respect to the norm are also point-wise close throughout the underlying set. This is an important property because functions in an RKHS are vectors in and of themselves, independent of the numerical value they take at points in their domain. Next, we present the definition of a very important class of functions.

\begin{definition*}
Let U be any set. A symmetric real-valued function $k: U \times U \rightarrow \mathbb{R}$ is said to be a positive definite kernel if for every $n \in \mathbb{N}$, we have that

\begin{equation}
    \sum_{i=1}^n \sum_{j=1}^n a_i a_j k(x_i,x_j) \geq 0
\end{equation}

\noindent
for any $n$ points $x_1,\,\cdots, x_n$ in $U$ and $n$ real numbers $a_1,\,\cdots, a_n$. 
\end{definition*}

The kernel matrix (also known as the \emph{Gram matrix}) $\mathcal{K}$ of a kernel $k$ with respect to a set of data points $\textbf{x}_1,\cdots,\textbf{x}_p$ in input space is a $p\times p$ matrix whose entries consists of the kernel evaluated at the associated data points, that is, the $ij$th entry contains the real number $k(\textbf{x}_i,\textbf{x}_j)$. This is a common way of storing the values of a kernel acting on a data set. The positive definite property of $k$ is equivalent to the Gram matrix $\mathcal{K}$ being \emph{positive semidefinite}, meaning that $A^T \mathcal{K} A \geq 0$ for every set of data points in the domain of $k$, and every $A \in \mathbb{R}^p$.

Another alternate characterization is that the reproducing kernel of an RKHS is a symmetric positive definite (p.d.)\ kernel. In the case that \emph{k} is the kernel of an RKHS, this property follows from the reproducing property and the fact that inner products are p.d.\ kernels. Conversely, every symmetric p.d. kernel induces a unique RKHS on its underlying set \citep[see][]{aronszajn1950theory}. This can be particularly useful for building new kernels out of old ones, as one can apply easier-to-prove properties of p.d.\ kernels to find new kernels. Some properties include: the sum (and product) of two p.d\. kernels is again a p.d.\ kernel, and a composition of a p.d.\ kernel with any function is again a p.d.\ kernel. A more extensive list can be found in \citet{theodoridis2015machine}, and the book \citet{scholkopf2002learning} contains an entire chapter dedicated to the design of kernels. The latter book also shows that new positive definite kernels can be constructed using any inner product space: If $\varphi$ is a function mapping a set $U$ into an inner product space, then $k(x,y) = \langle \varphi(x), \varphi(y) \rangle$ is a p.d.\ kernel. Finally, due to Mercer, we know that a continuous symmetric p.d.\ kernel has a series expansion of the form
\begin{equation}
\label{mercers theorem}
  k(x,y) = \sum_{i=1}^{\infty} \alpha_i^2 \varphi_i(x) \varphi_i(y)
\end{equation}

\noindent
where each $\alpha_i$ is a nonnegative real number, and each $\varphi_i$ is a real-valued function with some special properties. The details can be found in \cite{shawe2004kernel} or \cite{mercer1909xvi}, and the converse statement can be found in \cite{scholkopf2002learning}. This gives yet another way to check whether a given function is the kernel of an RKHS. We can write equation \ref{mercers theorem} more compactly as $k(x,y) = \varphi(x)^T \varphi(y)$, where $\varphi(x) = (\alpha_1\varphi_1(x), \,\alpha_2\varphi_2(x),\,\cdots)$. This convenient notation shows us that Mercer kernels act as a ``dot product" in the Hilbert space.

\smallskip

How does this connect back to statistical learning and more specifically to support vector machines? As a partial answer to the first question, there is a famous theorem known as the \emph{representer theorem} that very roughly states that any minimizing function of a minimization task can be represented by a finite linear combinations of functions $k_{\textbf{x}_i}$ where $k$ is the reproducing kernel belonging to the space of the minimizing function, and the $\textbf{x}_i$'s are training observations \citep{moguerza2006support}. This concept will come in handy later as we answer the second question.


\subsection{Using Kernels for Nonlinear Classification}
\label{Implementing Kernels in Nonlinear Classification section}

In more general applications, linear classification via hyperplane simply doesn't make for effective classification. If the decision boundary is highly nonlinear, it may not even be possible to find a hyperplane that divides the data in any meaningful way. When dealing with two dimensional data, a plot may reveal that a linear classifier cannot accurately capture where one class should end and the other should begin. Perhaps a parabola would result in a better fit if they are still separated from each other. What if one class is completely surrounded by another class? How do we come up with a classifier that can capture this geometry? First, we fix a kernel $k$ to build our support vector machine with. Selecting a kernel that minimizes the test error is a nontrivial task, and often many kernels will be used in the fitting process to find one that works best for the particular problem. We can then solve the same optimization problem as in the nonseparable case, but instead we replace the inner product in the dual representation of the problem with our choice of kernel $k$ (this is the so-called \emph{kernel trick}) \citep[see][ch.\ 11]{theodoridis2015machine}. Once again, the solution will yield a set of nonzero Lagrange multipliers associated to the constraints, and we can solve for $b$ in the same way as before. Thus for a choice of kernel $k$, the corresponding decision function $\hat{f}$ can be written as
\begin{equation}
\label{SVM prediction}
\hat{f}(\textbf{x}) = \sgn \left(\sum_{i\in \mathcal{S}} \lambda_i y_i k(\textbf{x}, \textbf{x}_i )+ b\right).    
\end{equation}

Throughout the process of solving the dual problem, as well as evaluating the decision function, we never actually need to perform any computations in the RKHS associated to $k$. All of the benefit brought by employing an RKHS can be accessed simply through evaluating $k$, which in many cases is a function of a familiar form. Some of the basic kernels used in practice are as follows. The first example is called the \emph{linear kernel}, and is given by the formula $k(\textbf{x},\textbf{y}) = \textbf{x}^{\,T} \textbf{y}$. This is simply the dot product in $\mathbb{R}^p$, and was the kernel used in section \ref{Hyperplane} to solve the problem of classifying observations using a linear decision boundary. It can readily generalize to the \emph{polynomial kernel}, given by $k(\textbf{x},\textbf{y}) = (\textbf{x}^{\,T} \textbf{y} + c)^d$, where \emph{c} and \emph{d} are hyperparameters that control the weight of lower-order terms and the degree of the polynomial, respectively. Naturally, the decision boundary that results from a polynomial kernel will take a very similar shape to the graph of a polynomial of the same degree. The Gaussian kernel, or \emph{radial basis function} (RBF) kernel is a very popular choice as it incorporates Euclidean distance, which in effect allows it to create highly nonlinear, ``oval-shaped" decision boundaries that bunch clusters of observations together. It is given by the formula $k(\textbf{x},\textbf{y}) = \exp{(-\lambda\| \textbf{x} - \textbf{y} \|^2)}$, where $\lambda > 0$ is a hyperparameter that controls the range of influence of the support vectors. It is common to see $\lambda$ parameterized with $1/(2\sigma^2)$, where $\sigma$ is taken as the tuning parameter. Interestingly, the feature space of the RBF kernel has an infinite number of dimensions. Quite often, the hyperparameters found in kernels are chosen via cross-validation (discussed in section \ref{Resampling}).

\smallskip

Through rigorous testing, SVMs have been found to offer good generalization performance \citep[ch.\ 11.10]{theodoridis2015machine}. What does this mean exactly? After fitting a model to a training set and tuning hyperparameters to reduce test error, one can bring in an entirely new data set not yet seen by the model, and find the \emph{generalization error}, which measures the accuracy of predictions. Empirical observations show that SVMs perform quite well in terms of having a low generalization error. While sparsification helps to reduce the computational requirement in practice, large data sets do pose a problem. There can still be many support vectors, and thus the computational load can still be high, but more importantly the training time scales poorly with larger data sets. For example, during the optimization process, an $n\times n$ matrix (called a \emph{Gram matrix}) is used for each possible comparison of predictor variables. Its size scales quadratically, and along with more technical details regarding the optimization problem, a major slow down in training time occurs for large data sets. \citet{hsu2003practical} provide a guide on good practices when performing support vector classification. One of the suggestions they make is to scale the entries of a Gram matrix before passing it to a function that fits an SVM. This can greatly improve runtime performance of the model fitting, as well as lead to accuracy improvements. Typically, one would linearly scale the training data to the unit interval, and then scale subsequent testing data using the exact same function. This technique is very important when dealing with \emph{graph kernels} (chapter \ref{chapter3}).


\subsection{SVMs for Multiclass Classification}

Although SVMs were not designed with multiclass classification in mind, there are ways that they may be implemented in this setting. We will cover the two most common approaches. The first approach is known as \emph{one-versus-one classification}. If we have $K > 2$ possible classifications, then for any two classes, the methods discussed in this chapter are used to construct an SVM solely based on these classes. This requires $\binom{K}{2}$ separate SVMs to cover all possible pairings of classes. Points are then classified according to which class they appear in the most across each SVM. The second approach is similar in spirit and is known as \emph{one-versus-all classification}. This time, for each fixed class, we treat the remaining $K-1$ classes as one large class, and fit an SVM between the two ``classes" as before. Points can then be classified based on which of the $K$ decision functions has the largest absolute value (in terms of equation \ref{SVM prediction}, we would simply ignore the sign function and take the absolute value of the inside function). As mentioned previously in the context of hyperplanes, the larger the magnitude of this inner function, the more confident we can be that the point has been classified correctly.


\chapter{Graph Kernels}
\label{chapter3}

The success of the kernel trick discussed in chapter \ref{chapter2} as a solution to nonlinear binary classification has sparked the investigation of more general applications of kernels in machine learning. In this chapter, we discuss the use of kernels to measure the similarity between \emph{graphs}, which are network-like structures consisting of vertices and edges that are used to model discrete data, such as molecules and compounds. We will mostly concern ourselves with various types of graph kernels and experimental results, and not necessarily be discussing them in conjunction with a learning method, although the experimental results found in the literature are usually obtained with a support vector machine.

\smallskip

In chemoinformatics, we are interested in the properties that chemical compounds have. As there are an enormous number of compounds, it is impossible to measure every property of every compound. Hence, it is useful to be able to predict properties that a chemical compound has based on knowledge of other compounds. This is done on the principal that similar compounds tend to have similar properties \citep{brown2009chemoinformatics}. Similarity of course has no clear cut definition, which forms the basis of research into graph kernels. As molecules are naturally represented as undirected labeled graphs, researchers come up with different measures that attempt to capture the similarities of these graphs as a way of quantifying how similar two molecules are to each other. Formulating these measure as positive definite kernels opens up the powerful \emph{kernel trick} for classification, which allows support vector machines to be nonlinear binary classifiers. This classification is how we infer similarity, and hence how properties may be predicted.

\smallskip

The way Graph Kernels are usually constructed is based on the framework introduced by \cite{haussler1999convolution}. Given two graphs $G_1$ and $G_2$, one can view them as the composition of subgraphs with certain properties, such as directed and undirected subgraphs, or paths, cycles, etc.\ \citep{ralaivola2005graph}. A graph kernel is then created from applying a positive-definite kernel on each pairing of subgraphs from $G_1$ and $G_2$, and taking the sum. Graph kernel designs are typically centred around one specific structural characteristic, dependent on what the application calls for. This is due to the fact that determining many graph properties without restriction are either NP-hard, NP-complete, or thought to be either NP-intermediate or NP-hard. For example, see \cite{gartner2003graph} for a proof that capturing all the information from a graph via subgraph isomorphism is NP-Hard. \cite{borgwardt2005shortest} also showed a similar result for the ``all-paths" kernel. In practice, this places a limit on the expressivity of graph kernels as similarity measures.

\smallskip

Section \ref{Introductory Concepts section} is dedicated to establishing the basic definitions and terminology regarding graphs, as well as an explanation of how kernels can use graph data to measure similarity. Section \ref{Baseline Graph Kernels section} details the initial graph kernels that began showing up in the early to mid 2000's that were popular or influential in the literature. Section \ref{More Advanced Examples section} contains descriptions of kernels that are slightly more complex and less general, but perform better than those of the previous section. Section \ref{Boosting Kernel Performance section} outlines techniques that have been developed to improve run time or classification accuracy of classes of kernels. Section \ref{Software and Data sets section} goes over common data sets and software implementations of graph kernels and support vector machines found in experiments in the literature. Finally, section \ref{Further Reading section} contains references to further readings, including more in depth treatments of the kernels discussed here, as well as kernels relevant to chemoinformatics that were not discussed here.


\section{Introductory Concepts}
\label{Introductory Concepts section}

\subsection{Essential Graph Theory}
\label{Essential Graph Theory section}

To begin, we review some basic definitions from graph theory. For more on the subject, see the standard textbook by \citet{diestel2017graphtheory}.

\begin{definition}
A graph $G = (V,E)$ consists of two sets; a vertex set $V = \{ v_1, \cdots, v_n\}$ containing $n$ vertices (or nodes) and an edge set $E \subset \{ (v_i,v_j) \in V \times V : v_i \neq v_j \}$ containing pairs of distinct vertices.
\end{definition}

Vertices are often thought of as representing the individual objects of interest, while each edge is thought of as a one-way connection between two objects. For example, we could represent a road network with a graph, where the vertices are intersections and the edges are roads. We could also model a molecule using a graph, taking the atoms to be vertices and the bonds between the atoms as edges. Unless otherwise stated, we shall assume that an arbitrary graph $G$ has a vertex set $V$ with $n$ vertices and an edge set $E$ with $m$ edges.

\smallskip

While intuitive, this representation cannot be readily implemented with typical operations and algorithms. Throughout subsequent sections, specific representations of graphs will be introduced as need. First, we look at a very important representation (even outside our scope of machine learning) known as the \emph{adjacency matrix}.

\begin{definition}
The adjacency matrix A of a graph $G$ is a $|V| \times |V|$ matrix where $[A]_{ij} = 1$ if $(v_i,v_j)$ is an edge in $G$ and 0 otherwise.
\end{definition}

This representation is particularly useful as the underlying graph may be examined and manipulated using well-understood techniques.

\begin{definition}
A graph $G$ is said to be undirected if for each edge $(v_i, v_j)\in E$, we have that $(v_j, v_i) \in E$.
\end{definition}

If the above property does not hold for every edge, we have what is known as a \emph{directed graph}. All chemical graphs will be undirected, however directed graphs are important for a specific class of graph known as \emph{trees} (section \ref{Tree-Pattern Graph Kernel section}). If a graph is undirected, then its adjacency matrix will be symmetric.

\begin{definition}
A walk of length m in a graph $G$ is a sequence of nodes $(v_i)_{i=1}^{m+1}$ where $(v_i,v_{i+1})$ is an edge in $G$ for every $i \in \{ 1, \,\cdots, m \}$. If $i \neq j$ implies that $v_i \neq v_j$, then we say that $(v_i)_{i=1}^{m+1}$ is a path. A walk with distinct vertices except for the start and end points is called a cycle.
\end{definition}

The existence of a walk between vertices implies a level of connection between them, depending on the context. For an undirected graph $G$, we say that $G$ is \emph{connected} if there exists a walk between any two vertices in $G$. Chemical graphs will always be connected. If $v$ is a vertex in $G$, the \emph{degree} of the vertex $v$, denoted by $\textup{deg}(v)$, is the number of neighbours of $v$, that is, the number of distinct walks of length one beginning at $v$. Finally, we have the notion of a \emph{labeled graphs} (for example, see \citet{mahe2004extensions}).

\begin{definition}
A labeled graph $G = (V,\,E,\,L)$ is a graph equipped with a labeling function $l: V \cup E \rightarrow A$, where $A$ is a set of labels.
\end{definition}

A graph's labeling function provides a way of differentiating both vertices and edges, and assigning properties. Graphs of molecules will always be labeled: the vertices represent atoms, and are labeled by their chemical symbol. The edges represent covalent bonds, and are sometimes labeled to indicate a type of covalent bond (single, double, etc.). In general, when referring to a graph $G$, we will assume it to be labeled, undirected, and connected, unless otherwise stated. The \emph{label sequence} $(l(v_1), l(v_1,v_2), \cdots, l(v_{m}, v_{m+1}), l(v_{m+1}))$ associated with a walk $(v_i)_{i=1}^{m+1}$ is a sequence containing the labels of every vertex and edge in the walk, in respective order \citep{mahe2004extensions}. For simplicity, we will sometimes refer to a label sequence associated with a walk as simply a labeled walk.


\subsection{Convolution kernels}
\label{Convolution substructure kernels section}

A very general construction of kernels on graphs is based on the framework of the \emph{convolution kernel}, introduced by \citet{haussler1999convolution}. We present the definition of the convolution kernel based on the form found in \citet{vishwanathan2010graph}.

\begin{definition}
Let $\mathcal{X}$ be a set of discrete objects, and let $x\in\mathcal{X}$ be an object with decomposition into $N$ components $\textup{\textbf{x}}_d = (x_1,\,\cdots, x_N)$. Define a function $R(x,\textup{\textbf{x}}_d)$ that outputs true if $\textup{\textbf{x}}_d$ is a valid decomposition of $x$, and false otherwise. If $R^{-1}(x)$ denotes the set of all valid decomposition's of a discrete object $x$, and $k_i$ is a kernel measuring the similarity between the ith components of $x$ for each $i\in \{ 1,\,\cdots, N\}$, then the convolution of the kernels $k_1,\,\cdots,k_N$ is defined as

\begin{equation}
    k(x,x') \defeq \sum_{\textup{\textbf{x}}_d \in R^{-1}(x)} \sum_{\textup{\textbf{x}}_d' \in R^{-1}(x')} \prod_{i=1}^N k_i (x_i,\,x'_i).
\end{equation}

\end{definition}

\begin{remark}
The feature map representation of the convolution kernel is given in \cite{kriege2014explicit}.
\end{remark}

\citet{haussler1999convolution} showed that this convolution is itself a positive definite kernel on $\mathcal{X}$, simply called a \emph{convolution kernel}. Many of the graph kernels encountered in the literature are instances of convolution kernels. For us, the sets $R^{-1}$ will contain decompositions of a graph into subgraphs \citep{rupp2010graph}. One may wish to consult the original paper of \citet{haussler1999convolution} for more technical and general details.

The kernels $k_i$ in the context of graph kernels often are used to compare vertices or edges of graphs, or their associated labels. The \emph{Dirac} (or Kronecker) kernel defined as $k(x_i,y_i) = I(x_i=y_i)$ \citep{rupp2010graph} is the most popular choice of base kernel in the literature, due to its efficiency and interpretability. See \citet{kriege2019unifying} for a description of its feature map.


\section{Baseline Graph Kernels}
\label{Baseline Graph Kernels section}

Now that we have a general understanding of how graph similarity can be quantified, we can discuss the first real attempts at constructing graph kernels.


\subsection{Simple Examples}
\label{simple examples of kernels section}

The convolution kernel is perhaps the most natural extension for positive definite kernel to discrete objects. In this short section, we present first-examples of the convolution kernel on graphs, as found in \citet{kriege2020survey}. They operate by comparing the labels between two graphs, while largely ignoring structure.

\begin{definition}
\label{VL kernel definition}
Let $G_1$ and $G_2$ be labeled graphs. The vertex label histogram (VH) kernel is defined as

\begin{equation}
K_{\textup{VH}} (G_1,G_2) \defeq \sum_{v_1\in V_1} \sum_{v_2 \in V_2} \textup{\textbf{I}}(l(v_1)=l(v_2)).
\end{equation}

\end{definition}

There is always a trade-off between \emph{expressivity} of a graph kernel \citep{kriege2020survey}, that is, the amount of structure and nuance it can take into account, versus the computational complexity. This kernel is one of the easiest to compute (having complexity $O(n^2)$), but treats graphs as a \emph{bag of components}, rather than a structured object. For example, two molecules are said to be isomers if they share the same number of atoms of each type, while possibly having different arrangements. This kernel would be unable to distinguish the graphs of a family of isomers. This is but one example illustrating the ``low-resolution" of the VH kernel. While not a focus of this review, the computational limits of graph kernels generally are discussed in some papers such as \citet{gartner2003graph}, \citet{borgwardt2005shortest}, and \citet{kriege2020survey}.

\begin{definition}
\label{EL kernel definition}
Let $G_1$ and $G_2$ be labeled graphs. The edge label histogram (EH) kernel is defined as

\begin{equation}
k_{\textup{EH}} (G_1, G_2)\defeq \sum_{(v_1,v_1')\in E_1} \sum_{(v_2,v_2')\in E_2} \textup{\textbf{I}}[l(v_1)=l(v_2)]\cdot \textup{\textbf{I}}[l(v_1,v_1')=l(v_2,v_2')] \cdot \textup{\textbf{I}}[l(v_1')=l(v_2')].
\end{equation}

\end{definition}

It is clear that the EH kernel has complexity $O(m^2)$. The corresponding feature maps for both of these kernels can be found in \citet{sugiyama2015halting}. These kernels are not very accurate as is. The experiments in the literature that include them typically do so for a lower-bound on performance. However, as noted by \citet{sugiyama2015halting}, they are especially important to measure against random walk kernels such as the geometric random walk kernel, as a process known as halting can make those much more complex kernels perform similarly to these ``trivial" kernels. See section \ref{Direct Product Kernel section} for more details. These simple kernels can be surprisingly effective with slight modification; see section \ref{Graph RBF Kernel Section} for a brief discussion on composing linear kernels with the RBF kernel.


\subsection{Direct Product Kernel}
\label{Direct Product Kernel section}

The first graph kernels began appearing in the literature in 2003, with papers by \citet{gartner2003graph} and \citet{kashima2003marginalized} \citep{kriege2020survey}. These kernels measure graph similarity by comparing all possible pairings of labeled walks between two graphs. This section is focused on the \emph{direct product kernel} of \citet{gartner2003graph} and its variations. We postpone the discussion of the marginalized kernel of \citet{kashima2003marginalized} until section \ref{Marginalized Kernels}. Let us begin with the definition of the \emph{direct product graph} (see \citet{gartner2003graph}). We use the notation from \citet{vishwanathan2010graph}.

\begin{definition}
\label{Direct Product Graph}
Let $G_1,\,G_2$ be two labeled graphs. The direct product graph $G_{\times}$ of $G_1$ and $G_2$ is a graph with vertex set
\smallskip
\[
V_{\times} \defeq \{ (v_1, v_2) \in V_1\times V_2 : l(v_1) = l(v_2)\}
\]

\noindent
and edge set
\smallskip
\[
E_{\times} \defeq \{ \left((v_1, v_2),(v_1', v_2')\right) \in V_{\times}\times V_{\times} : (v_1, v_2) \in E_1,\, (v_1', v_2') \in E_2,\, l((v_1, v_2)) = l(v_1', v_2')\}.
\]

The graphs $G_1$ and $G_2$ are called the factor graphs of $G_{\times}$.
\end{definition}

\begin{remark}
Note that this formulation of the direct product graph requires vertices and edges from the the factor graphs to have matching labels. Graph-theoretic applications involving direct product graphs may omit this restriction from the definition.
\end{remark}

The direct product graph is useful as it allows for simultaneous labeled-walk comparison on the factor graphs. It can be shown that the number of walks with a particular label sequence on $G_{\times}$ is exactly the product of the number of walks with that label sequence in each factor graph $G_1$ and $G_2$ \citep[Prop.\ 3]{gartner2003graph}.

\smallskip

For simplicity, we first present a special case of the direct product kernel called the \emph{vertex edge label histogram kernel} (VEH), which was introduced by \citet{sugiyama2015halting}. It combines both the vertex (definition \ref{VL kernel definition}) and edge (definition \ref{EL kernel definition}) label histogram kernels; however, the direct product adjacency matrix representation \citep[Lemma 1]{sugiyama2015halting} makes its function rule more compact and helps illustrate its connection with the other direct product kernels. The feature map representation can be found in the original paper.

\begin{definition}
\label{vertex-edge label kernel definition}
Let $G_1$ and $G_2$ be two labeled graphs, with $G_{\times}$ denoting their direct product. The vertex edge label histogram kernel is defined as

\begin{equation}
    k_{\textup{VEH}} (G_1, G_2) \defeq \sum_{i=1}^{|V_{\times}|} \sum_{j=1}^{|V_{\times}|} \left[ A_{\times} \right]_{ij}
\end{equation}

\noindent
where $A_{\times}$ is the adjacency matrix of $G_{\times}$.
\end{definition}

This kernel, counting the entries of $A_{\times}$, in effect counts the number of edges in both graphs with matching edge labels and associated vertex labels, or equivalently, counting random labeled walks of length one.

\smallskip

Next, we present the general direct product graph kernel of \citet{gartner2003graph}. It takes into account \emph{every} labeled walk of two given graphs.

\begin{definition}
\label{direct product kernel definition}
Let $G_1$ and $G_2$ be two graphs, and $G_{\times}$ their direct product graph with adjacency matrix $A_{\times}$. Let $(\lambda_n)_{n=0}^{\infty}$ be a sequence of weights consisting of nonnegative real numbers with the property that the matrix power series

\begin{equation}
    \sum_{n=0}^{\infty} \lambda_n A_{\times}^n
\end{equation}

\noindent
converges to a matrix $M$. Then the direct product kernel $k_{\times}$ is defined as the entry-wise sum of the matrix $M$:

\begin{equation}
k_{\times}(G_1, G_2) \defeq\, \sum_{i=1}^{|V_{\times}|} \sum_{j=1}^{|V_{\times}|} \left[ M \right]_{ij} .
\end{equation}

\end{definition}

\begin{remark}
We avoid discussing the precise notion of convergence of sequences and series of matrices, instead mentioning a couple of common choices for the weight sequence that will ensure the matrix power series converges and has a familiar closed-form. See \citet{horn2012matrix} for details regarding power series of matrices.
\end{remark}

The kernel uses the adjacency matrix representation of graphs, as it can generate the number of walks between vertices in a graph. For a graph $G$ with adjacency matrix $A$, the $ij$th element of $A^k$ contains the number of walks of length $k$ from vertex $v_i$ to $v_j$ (see \citet{duncan2004powers}, for example). Thus for a direct product graph $G_{\times}$ and $\textbf{v}_i = (v_1,v_2),\, \textbf{v}_j = (v_1',v_2') \in V_{\times}$, the $ij$th entry of $A_{\times}^k$ contains the number of walks of length $k$ between $v_1$ and $v_1'$, and $v_2$ and $v_2'$ that have the same label sequence. 

\citet{gartner2003graph} present two examples of weight sequences with the desired convergence property. The first corresponds to the coefficients of the power series of the exponential function, which are given by $\lambda_n = \frac{1}{n!}$. The advantage is that the resulting matrix power series converges for any square matrix with real entries \citep{horn2012matrix}. However, due to the nature of matrix multiplication, the exact limit may be difficult to compute for general matrices. For diagonalizable matrices, there is a known closed-form \citep{gartner2003graph}. Otherwise, one can sum the first $N$ terms to estimate the kernel value. The second sequence\textemdash more widely seen in the literature\textemdash chooses the weights so that the series converges like a geometric series. More specifically, let $\gamma \in (0, \frac{1}{a})$, where $a \geq \max_{v\in V_{\times}} \text{deg}(v)$, and set $\lambda_n = \gamma^n$. Then it can be shown that the matrix power series converges to

\begin{equation}
    \sum_{n=0}^{\infty} \left(\gamma A_{\times}\right)^n = (I - \gamma A_{\times})^{-1}
\end{equation}

\noindent
where $I$ is the identity matrix. The corresponding kernel is called the \emph{geometric random walk kernel} (GRW) \citep{sugiyama2015halting}:

\begin{equation}
    k_{\textup{GRW}}(G_1, G_2) \defeq\, \sum_{i=1}^{|V_{\times}|} \sum_{j=1}^{|V_{\times}|} \left[ (I - \gamma A_{\times})^{-1} \right]_{ij}.
\end{equation}

In this case, the computational complexity is dictated by the complexity of the matrix inversion algorithm, which are approximately $O(n^3)$ for an $n\times n$ matrix. In the worst-case, the dimension of $G_{\times}$ is $n^2\times n^2$ where $n$ is the number of nodes in both factor graphs, hence the complexity of computing the direct product kernel is $O(n^6)$ \citep{borgwardt2020graph}. This is quite slow, but has been drastically improved by \citet{vishwanathan2010graph}, who were able to achieve a complexity of $O(dn^3)$ for this kernel, where $d$ is the total number of labels, using fixed-point iterations and conjugate gradient methods. This was used in an experiment by \citet{sugiyama2015halting} with labeled graphs, which we discuss below. Furthermore, \citet{kang2012fast} introduced methods to approximate the kernel that allows for further reduction in complexity.

\smallskip

While \citet{gartner2003graph} provide no experimental results alongside their direct product kernel, there are other instances in the literature where it has been tested. For example, \citet{sugiyama2015halting} studied the problem of \emph{halting} found in random walk kernels, where the weights associated with longer walks are so small that walks of length one dominate the kernel value. This issue stems from the choice of coefficients that ensure convergence, which approach zero rather quickly for the geometric and exponential variants. To test halting, the authors introduce the \emph{N-step random walk kernel}, which modifies the direct product kernel by truncating the infinite series after a finite number of terms, meaning only finite walks are incorporated.

\begin{definition}
\label{N-step random walk kernel definition}
Let $G_1$ and $G_2$ be two labeled graphs, with $G_{\times}$ denoting their direct product, and $N\in\mathbb{N}$. The N-step random walk kernel is defined as

\begin{equation}
    k_{\times}^N (G_1, G_2) \defeq \sum_{i=1}^{|V_{\times}|} \sum_{j=1}^{|V_{\times}|} \sum_{n=0}^N \lambda_n \left[ A_{\times}^n\right]_{ij}
\end{equation}

\noindent
for some real coefficients $\lambda_n,\, n\in \{ 0,\,\cdots,N\}$.
\end{definition}

With convergence no longer an issue, a more natural weighting scheme could be applied, making the kernel more flexible. For their experiments, the weights of this kernel were all set to 1 and the number of steps $N$ range from 1 to 10. Other kernels were used in the experiments as well. The VEH kernel was used as a point of comparison against the GRW kernel to test their theoretical results on halting, and the \emph{Weifeiler-Lehman subtree} (WLS) kernel (see section \ref{WL kernels}) was used, as it is known to be one of the most consistently accurate graph kernels. The data sets used are the standard ones found in the literature: ENZYMES, NCI1, NCI109, MUTAG, and D\&D, which all consist of labeled nodes \citep{shervashidze2011weisfeiler}. While the WLS kernel came out on top in terms of accuracy of predictions, the $N$-step random walk kernel with an optimal choice of steps (4-6 steps in this experiment) outperformed the GRW kernel, sometimes by up to 5\%. The effect of halting was amplified when the parameter in the GRW kernel was set much smaller than its theoretical maximum. The higher the parameter is set, the more likely convergence will become a problem should a previously unseen pair of graphs create a direct product graph with a larger maximum degree. If one wants to implement the GRW kernel, \citet{sugiyama2015halting} recommends including the simple vertex and edge label histogram kernels (see section \ref{simple examples of kernels section}) in the initial testing as a baseline to measure performance against.


\subsection{Shortest-Path Graph Kernel}
\label{Shortest-path Graph Kernel section}

The next big advancement in the graph kernel literature was the \emph{shortest-path kernel}, introduced by \citet{borgwardt2005shortest}. One of the main goals for the authors was to improve runtime over the other kernels available at the time that suffered from long computation times, such as the direct product kernel discussed in the previous section.

\smallskip

First, we provide the definition of an edge walk from \citet{borgwardt2005shortest} that is very similar to that of a walk, but instead emphasises the edges rather that the vertices. For a graph $G$, an \emph{edge walk} is a sequence of edges $(e_i)_{i=1}^m$ in $E$ with the property that for each $1<i\leq m$, $e_{i-1} = (v_{i-1},v_{i-1}')$ and $e_{i} = (v_{i},v_{i}')$, we have $v_{i-1}'=v_i$. The shortest-path graph kernel uses the adjacency matrix representation of graphs. However, it does so after what is known as the \emph{Floyd-transformation} is applied to the graphs. This transformation can be done using the \emph{Floyd-Warshall} algorithm \citep{floyd1962algorithm}, which takes as input the adjacency matrix of a graph and a matrix containing the weights of edges (or distance between vertices). The output is a new matrix, where the $ij$th element contains the length of the shortest path between the $i$th and $j$th vertex. This new matrix is an adjacency matrix for the transformed graph. If $n$ denotes the number of vertices in the original graph, then the runtime of this algorithm is $O(n^3)$. For chemical graphs, one may use the adjacency matrix as the weight matrix. Pseudo-code for the algorithm can be found in the paper by \citet{borgwardt2005shortest}. We are now able to present the general form of the \emph{shortest path kernel}, which is yet another instance of the convolution kernel framework.

\begin{definition}
\label{shortest path kernel definition}
Let $G_1$ and $G_2$ be two graphs, with corresponding Floyd-transformed graphs $F_1$ and $F_2$. The shortest-path graph kernel is defined as

\begin{equation}
\label{shortest path kernel equation}
    k_{\textup{SP}}(G_1,G_2) \defeq \sum_{e_1 \in E_{F_1}}  \sum_{e_2 \in E_{F_2}} k_{walk}^1 (e_1, e_2)
\end{equation}

\noindent
where $k_{\textup{walk}}^1$ is a positive definite kernel on edge walks of length 1.
\end{definition}

\begin{remark}
The feature map of $k_{SP}$ can be found in \citet{shervashidze2011weisfeiler}.
\end{remark}

As the number of edges in the Floyd-transformed graph is $n^2$, where $n$ is the number of vertices in the original graph, it follows from equation \ref{shortest path kernel equation} that the shortest-path kernel has a computational complexity of $O(n^4)$. An immediate improvement over many standard forms of random walk kernels is the built-in prevention of \emph{tottering}, which helps eliminate noise. Tottering occurs in a walk when a vertex is immediately returned to after stepping away from it, and tends to record uninformative features. Typically, the base kernel $k_{walk}^1$ is the product of three kernels. If $e_1 = (v_1,v_1')$, $e_2 = (v_2, v_2')$, then 

\begin{equation}
\label{1-walk form equation}
    k_{\textup{walk}}^1(e_1, e_2) = k_l (l(v_1),l(v_2)) \cdot k_d (d(v_1,v_1'),d(v_1,v_1')) \cdot k_l (l(v_1),l(v_2))
\end{equation}

\noindent
where $k_l$ is a kernel on vertex labels, such as the Dirac kernel, and $k_d$ is a kernel defined on the shortest-path distance. When analyzing chemical graphs, a default distance function is used, where the distance between two vertices is the number of edges in the shortest path between them. Moreover, chemical graphs are usually connected, thus every pair of vertices can be assigned a distance. In experiments, $k_d$ is also taken to be the Dirac kernel.

\smallskip

For testing, \citet{borgwardt2005shortest} used 10-fold one-versus-all SVM classification on a bioinformatic data set containing 540 proteins. They elected to use the Dirac kernel on vertex labels, but used the \emph{Brownian bridge kernel} on both edge and vertex length (adding another kernel to the product in equation \ref{1-walk form equation}). Its definition has been omitted in this review as weighted graphs are rarely found in chemoinformatics. The test results showed a clear improvement over random walk-kernels, both in terms of runtime and classification accuracy. The shortest-path kernel has been used in many other tests as well. \citet{shervashidze2011weisfeiler} compared the shortest-walk kernel to variants of the \emph{Weisfeiler-Lehman} graph kernels (section \ref{WL kernels}). Their experiments showed that the shortest-path kernel outperformed many of the kernels that came before it by a large margin, including variations of the random walk kernel. Now, the Weisfeiler-Lehman techniques introduced can be applied to many different kernels, as it functions as a label refinement algorithm. Applying this to the shortest-path kernel, they found that it resulted in the most accurate classifier, while remaining quick to compute on most data sets. 


\section{Specialized Examples}
\label{More Advanced Examples section}

The kernels in the last section are standard in the wider field of graph kernels. In this section, we will look at graph kernels the were designed with chemical graphs and chemoinformatic applications in mind. It is quite common to find the kernels of section \ref{Baseline Graph Kernels section} in software packages for graph kernels. More specialized kernels are less readily-available in general, however each kernel in this section can be computed using the freely-available \href{http://chemcpp.sourceforge.net/html/index.html}{ChemCPP} package.


\subsection{Marginalized Kernels}
\label{Marginalized Kernels}

Many iterations of marginalized kernels have appeared throughout the literature. First introduced by \citet{kashima2003marginalized}, these random walk kernels offer more control over the direct product kernel by having user-set probabilities for each labeled walk. We follow both \citet{mahe2004extensions} and \citet{mahe2005graph}, which extend the original formulation by \citet{kashima2003marginalized} with the goal of achieving better results on chemical graphs. \citet{vishwanathan2010graph} provided a generalization of this kernel, along with algorithms that improve computational complexity. Unfortunately, vertex labels are not addressed, which is not particularly useful in traditional chemoinformatic applications. The computational improvements have still been applied to labeled graphs, as discussed in section \ref{Direct Product Kernel section}.

\smallskip

Let $G$ be a labeled graph. The set $V^* = \cup_{n \in \mathbb{N}} V^n$ is used to denote the set of finite-length sequences of vertices. If $\textbf{v} \in V^*$, then $l(\textbf{v})$ is used to denote the associated label sequence, which was defined in section \ref{Essential Graph Theory section}. The general form of the marginalized graph kernel of \citet{kashima2003marginalized} is given in the following definition from \citet{mahe2004extensions}.

\begin{definition}
Let $G_1$ and $G_2$ be two labeled graphs. The marginalized graph kernel is defined as

\begin{equation}
\label{marginalized graph kernel}
    k_{\textup{M}}(G_1,G_2) \defeq \sum_{h_1\in V_1^*} \sum_{h_2\in V_2^*} p_1(h_1) p_2(h_2) k_L(l(h_1),l(h_2)) .
\end{equation}

\noindent
where $p_1$ and $p_2$ are probability distribution on $V_1^*$ and $V_2^*$, respectively, and $k_L$ is a kernel defined on label sequences.
\end{definition}

One could simply use the Dirac kernel as a means of comparing label sequences. \citet{mahe2004extensions} provide a brief explanation on how to compute this kernel, which makes use of product graphs and geometric series, as well as how to set the probability distributions. The authors then present two ways of modifying the kernel in equation \ref{marginalized graph kernel}. First is an iterative vertex label transformation called the \emph{Morgan index} \citep{morgan1965generation} process that seeks to highlight paths containing more relevant information, improving performance and cutting down computational time. The process is as follows: to start, each vertex is labeled with ``1". For each iteration $i$ in some predefined range and each atom $A$ in the molecule, the label of $A$ is increased by the sum of the label values of its direct neighbours from the previous iteration. To compute this in practice, let $M_0$ be the vector of dimension equal to the number of atoms, and populate $M_0$ with ones to represent the labels. On the $i$th iteration, the vector defined by $M_i = AM_{i-1} = A^i M_0$, where $A$ is the adjacency matrix of the graph, contains the desired labels. The Morgan index has also been used with the \emph{treelet kernel} \citep{gauzere2012two}.

The second way to modify the marginalized graph kernel is to change its underlying probabilistic model. This is done to prevent the phenomenon of \emph{tottering}, which occurs when a random walk revisits a vertex immediately after stepping away from it. Walks containing totters are thought to introduce unwanted noise when attempting to learn a model. This is theoretically implemented using a 2nd-order Markov model; however, in practice a graph transformation is applied to the two graphs which eliminates the possibility of tottering walks, meaning that the kernel ends up using the same 1st-order Markov process as the original formulation, except on a larger transformed graph with an increased complexity (from $O(n^2)$ to $O((n+m)^2)$). Given two graphs $G_1$ and $G_2$ and their corresponding transformed graphs $G_1'$ and $G_2'$, respectively, the kernel can be written as

\begin{equation}
    k_{\textup{MNT}}(G_1,G_2) \defeq \sum_{h_1'\in (V_1')^*} \sum_{h_2'\in (V_2')^*} p_1'(h_1') p_2'(h_2') k_L(l'(h_1'),l'(h_2')).
\end{equation}

The paper provides experimental results for these kernels using SVMs, which were implemented using \href{https://svm.msl.ubc.ca/gist/}{GIST}. The first data set used was the MUTAG data set. The use of 1st and 2nd order Markov random walk models had a very small impact on classification ability. On the high end, less than 2\% in performance was gained when the 2nd order model was used to eliminate tottered walks. Using the Morgan index iteration on the graph vertices, a small increase in classification was generally observed when between one and three iterations were performed. Applying these iterations did lead to a dramatic decrease in computational time. Two iterations alone reduced computational time by a factor of approximately 240. The second data set used was the PTC data set. The best results were obtained after between 8 and 10 Morgan index iterations, peaking at $\sim\!63$ ROC area (for context, we would expect about 50 ROC area if we randomly classified data).

The ChemCPP toolbox contains functions to compute both the marginalized graph kernels of \citet{kashima2003marginalized} and the extended marginalized kernel of \citet{mahe2005graph}.


\subsection{Tree-Pattern Graph Kernel}
\label{Tree-Pattern Graph Kernel section}

Another popular substructure used to compare graphs is a special type of subgraph known as a \emph{tree}. There is a notion in which the structure of organic molecules resemble that of a tree (\citet{yamaguchi2003graph}, \citet{kriege2020survey}), and based on empirical evidence, it appears that classifying organic molecules based on trees embedded within the graph may lead to higher accuracy. In the paper by \citet{mahe2009graph} which this section follows, a graph kernel is proposed that can measure similarity based on common subtrees, with a parameter controlling tree complexity. This builds off of the work of \citet{ramon2003expressivity} with the hope that the kernel provides a way to capture physicochemical properties of atoms. \citet{mahe2009graph} also provide recursive algorithms to compute their kernels in practice, which we omit from this presentation.

\smallskip

We begin by giving the appropriate definitions related to trees from \citet{mahe2009graph}. In this context, there are two important sets associated with each vertex $v\in V$ in a directed graph $G$. The set of incoming neighbours $\delta^-(v) = \{u\in V : (u,v)\in E \}$, and the set of outgoing neighbours $\delta^+(v)=\{ u\in V : (v,u)\in E \}$. The in-degree of a vertex $v$ is the quantity $\labs \delta^-(v) \rabs$ and the out-degree is the quantity $\labs \delta^+(v) \rabs$. A \emph{rooted tree} $t$ is a directed, connected graph containing no cycles where every node has in-degree 1, except for one node having in-degree 0, which is known as the \emph{root} of the tree. For the remainder of this section, all trees will be rooted trees. The nodes of $t$ with out-degree 0 are leaf nodes, while the remaining nodes will be called interior nodes. The depth of a node is defined as the length of the path from the root to it plus one. A tree where each leaf node has the same depth $n$ is called a perfectly depth balanced tree of order $n$, or simply a \emph{balanced tree}.

\smallskip

To define the tree-pattern graph kernel, a way of formalizing the notion of a tree being embedded in a graph is needed, as well as a way to count how many times a tree is found within the graph. The following two definitions accomplish this.

\begin{definition}
Let $G = (V,\,E)$ be a graph and let $t = (V_t,\,E_t)$ be a tree with vertex set $V_t = \{ \tau_1,\,\cdots, \tau_N \}$. We say that a $n$-tuple of vertices $(v_1,\, \cdots,\, v_n) \in V^n$ is a tree-pattern of $G$ with respect to $t$ if the following properties hold:

\begin{enumerate}[i]
    \item\!\!\!) Matching vertex labels: For every $i\in \{ 1,\,\cdots, n\}$, $l(v_i) = l(\tau_i)$.
    \item\!\!\!) Edges correspondence: For every $(\tau_i,\tau_j) \in E_t$, we have that $(v_i,\,v_j)\in E$ and $l((v_i,v_j)) = l(\tau_i,\tau_j))$. Moreover, if we also have that $(\tau_i,\tau_k) \in E_t$, then $j\neq k$ if and only if $v_j \neq v_k$.
\end{enumerate}

The set of all such $n$-tuples will be denoted by $\textup{Pattern}(t,G)$.
\end{definition}

An important subtlety of this definition is that a vertex in the graph may be used multiple times in the tree pattern. In effect, this will allow for two edges with opposite orientation connecting two vertices (i.e.\ the edges in chemical graphs, which represent covalent bonds) to be part of the tree-pattern.

\begin{definition}
The tree-pattern counting function, denoted by $\varphi_t(G) = | \textup{Pattern}(t,G) |$, counts the number of occurrences of a tree pattern $t$ in $G$.
\end{definition}

We may now present the general form of the tree-pattern graph kernel, which is another instance of the convolution kernel framework (\citet{rupp2010graph}, \citet{vishwanathan2010graph}).

\begin{definition}
\label{Tree-pattern Kernel defintion}
Let $T$ be a set of trees, $w(t)$ a nonnegative weight function defined on $T$, and $G_1$, $G_2$ two labeled graphs. The tree-pattern graph kernel is defined as

\begin{equation}
\label{Tree-pattern Kernel eq}
    k_{\textup{TPK}}(G_1,G_2) \defeq \sum_{t\in T} w(t) \varphi_t(G_1) \varphi_t(G_2)
\end{equation}

\noindent
where $\varphi_t$ is the tree-pattern counting function.

\end{definition}

The two specific forms of tree-pattern graph kernels found in this paper both take $T$ to be the set of balanced trees of order $h$, which they denote by $B_h$. Where the two kernels differ is how they assign weight to each tree. Let $\lambda \geq 0$ be a nonnegative hyperparameter. The \emph{size-based balanced tree-pattern kernel} $k_{\text{size}}^h$ is equation \ref{Tree-pattern Kernel eq} with $w(t) = \lambda^{|t|- h}$, where $|t|$ denotes the number of nodes in the tree:

\begin{equation}
    k_{\text{size}}^h(G_1,G_2) \defeq \sum_{t\in B_h} \lambda^{|t|- h} \varphi_t(G_1) \varphi_t(G_2).
\end{equation}

This kernel may be generalized by enlarging the set of trees under consideration, such as using the set of trees of depth up to and including $h$, denoted by $T_h$. The \emph{branching-based balanced tree-pattern kernel} $k_{\text{Branch}}^h$ is equation \ref{Tree-pattern Kernel eq} with $w(t) = \lambda^{\textup{branch}(t)}$, where $\textup{branch}(t)$ equals the number of leaf nodes in $t$ plus one:

\begin{equation}
    k_{\text{branch}}^h(G_1,G_2) \defeq \sum_{t\in B_h} \lambda^{\textup{branch}(t)} \varphi_t(G_1) \varphi_t(G_2).
\end{equation}

A goal of the authors was to control the complexity of tree patterns incorporated into the kernels. Informally, higher complexity of a tree means it has a high number of leaf nodes, or a high number of internal nodes. Thus complexity refers to how ``nonlinear" a tree is. When $\lambda > 1$, more weight is placed of complex tree patterns, and when $\lambda < 1$, more weight is put on simpler tree patterns \citep{rupp2010graph}. In fact, the authors remark that as $\lambda$ approaches 0, both kernels approach the value of the direct product kernel (section \ref{Direct Product Kernel section}). The computational complexity of both of these tree-pattern kernels is $O(h n^2  d^{2d})$, where $d$ is the upper bound on the number of out-degrees of the vertices and $n$ is the number of vertices in one graph. For chemical compounds in chemoinformatics, \citet{mahe2009graph} claim that it is almost always the case that $d$ is no greater than 4. Moreover, due to the computational algorithms introduced, the extension of the size-based kernel to the set $T_h$ carries no extra computational cost.

\smallskip

A phenomenon known as \emph{tottering}, first noticed with walk-based kernels \citep{mahe2005graph}, can also impact the effectiveness of these tree kernels. Tottering happens when a tree-pattern contains a vertex from the graph as both the parent and child of another vertex. Tree-patterns of this type end up adding noise that tends to obscure the important features, as they vastly outnumber non-tottering tree patterns as the depth of the tree considered increases. The way the authors ultimately handle this is to apply a transformation to the graph that produces a new graph without any tottering walks (the same transformation mentioned in section \ref{Marginalized Kernels}). This \emph{non-tottering kernel} takes the form

\begin{equation}
    k_{\text{NT}}(G_1,G_2) = \sum_{t\in T} w(t) \varphi_t^{NT}(G_1') \varphi_t^{NT}(G_2')
\end{equation}

\noindent
where $\varphi_t^{NT}$ is the no-tottering tree-pattern counting function (see the original paper), and $G_1',\,G_2'$ are the transformed graphs of $G_1,\,G_2$, respectively. This ends up scaling the computational complexity by

\begin{equation}
\label{complexity scaling factor}
    \frac{(n + m)^2}{n^2}
\end{equation}

\noindent
and hence the resulting complexity is $O(h ((n+m) d^{d})^2)$. The average observed value of equation \ref{complexity scaling factor} across the experimental data sets used by the authors was found to be approximately equal to 9.

\smallskip

\citet{mahe2009graph} provide experimental results on their proposed kernels and their extensions. Classification was performed with SVMs using \href{https://www.csie.ntu.edu.tw/~cjlin/libsvm/}{LIBSVM}, as well as the \href{http://pyml.sourceforge.net/}{PyML} framework. The graph kernels were computed using \href{http://chemcpp.sourceforge.net/html/index.html}{ChemCPP}. The first series of experiments were performed on two public data sets, the first being the common MUTAG data set and the second being the now unavailable NCI GI50 data set of potential tumor-suppressing compounds. Optimal values of $\lambda$ appeared to decrease for larger $h$, and the author's reasoning was that the number of tree patterns increases exponentially as $h$ increases and a smaller $\lambda$ allows for less individual influence of each tree. When $h$ got too large ($h\geq8$), they noticed convergence issues if $\lambda$ wasn't small enough. However, when testing the no-tottering extensions, this problem went away and a small accuracy improvement was seen. It appears that the optimal value of $h$ is highly dependent on the data set and what  differentiate the molecules. Overall, the kernels performed better than the walk-based kernels. In \citet{sawada2014benchmarking}, these kernels are tested against a few other common kernels. They are mostly comparable to Tanimoto kernels, but ultimately lose out against them in nearly every test when it comes to accuracy.


\subsection{Tanimoto Kernels}
\label{Tanimoto Kernels Section}

In the paper by \citet{ralaivola2005graph}, three closely related kernels are developed for direct application to problems in chemoinformatics. These kernels represent molecules using a technique known as \emph{molecular fingerprinting}, which encodes certain paths\footnote{The authors use a weaker definition of path than what we presented in section \ref{Essential Graph Theory section}, only requiring paths to have distinct edges, and not necessarily distinct vertices. We will keep consistent with this terminology only in this section.} found within the graph in feature vectors. The thinking is that the molecular fingerprinting representation is more apt for classifying the molecules found in organic chemistry. The way in which molecules are compared is based on the \emph{Tanimoto similarity measure} \citep{fligner2002modification}, which is a well-known measure of chemical distance. 

\smallskip

Viewing molecules as labeled graphs, the molecular fingerprinting technique assigns values to paths emanating from each vertex, and stores them in a feature vector. The paths are computed via a depth-first search algorithm, for which not all implementations are created equal (see \citet{ralaivola2005graph} for descriptions of common implementations). A decision needs to be made on whether cycles are allowed, the maximum depth, and if the same edge can be visited in two paths with the same starting node after the first point of divergence. For an example of the latter, if \emph{a}, \emph{b}, \emph{c}, and \emph{d} label a square-shaped graph in the clockwise sense, and if the path $(a,b,c,d)$ is traversed first, then $(a,d,c)$ would not be traversed. All of these modifications have an effect on both complexity and effectiveness of fingerprinting. For example, if a molecule contains $n$ atoms and $m$ bonds, then finding all paths up to depth $d$ with the edge-divergence condition is $O(nd)$. Without the edge-divergence condition, the complexity becomes $O(n\alpha^d)$, where $\alpha$ is the average number of atoms that are neighbours to any given atom in the graph. The authors note that the value of $\alpha$ is typically low for organic molecules.

\smallskip

Let $G$ denote the graph of a molecule, and let $\mathcal{P}(d)$ denote the set of all atom-bond labeled paths of length $d$ in $G$. There are two approaches to fingerprinting that are considered by \citet{ralaivola2005graph}. The first is called the \emph{binary feature map of $G$ for a depth $d$}, and is a vector-valued function given by

\begin{equation}
    \varphi_d(G) = (\textbf{I}(p\subset G))_{p\in\mathcal{P}(d)}
\end{equation}

\noindent
where $\textbf{I}(p\subset G)$ is the indicator function that outputs 1 if at least one depth-first search on $G$ of depth at most $d$ produces the path $p$, and 0 otherwise. Notice that this disregards multiple separate instances of $p$ appearing in $G$. The second approach is called the \emph{counting feature map}, denoted by $\phi_d$, and is given by

\begin{equation}
    \phi_d (G) = (\#\{p\subset G\})_{p\in\mathcal{P}(d)}.
\end{equation}

This feature map does record the number of times a path $p$ appears in the graph $G$. The notation $\#\{p\subset G\}$ is adapted from \citet{Klambauer2015}. The third approach to fingerprinting is a fixed-length binary feature map of depth $d$ and length $r$, denoted by $\overline{\varphi}_{d,r}(G)$. Every $p\in\mathcal{P}(d)$ is mapped to $b$ indices between 1 and r by some function $f$. The corresponding indices in the vector $\overline{\varphi}_{d,r}(G)$ are then set to 1 (and remain 1 even if it is mapped to again). Standard implementations for $f$ fix $b$ to be either 1 or 4, and compute a hash value for each path $p$ as a seed of a random number generator that chooses $b$ random integers, and reduces them mod $r$. Note that one may wish to use very large length feature vectors, having a bit position for each possible path, to eliminate the possibility of hash collisions that can cause loss of information.

The kernels introduced in this paper are built out of the simpler dot product kernels on the molecular fingerprint vectors. For example,

\begin{equation}
\label{dot product of binary vectors}
    k_{\varphi_d}(G_1,G_2) \defeq \langle \varphi_d(G_1),\,\varphi_d(G_2) \rangle = \sum_{p\in\mathcal{P}(d)} \textbf{I}(p\subset G_1) \textbf{I}(p\subset G_2).
\end{equation}

When using the binary feature map of $G$ of size $r$, this dot product kernel is denoted by $k_{d,r}$. The dot product kernel using the counting feature map is not explicitly used. We are now ready to present the main kernel of this section.

\begin{definition}
Let $G_1,\,G_2$ be two graphs, and $d\in\mathbb{N}$ denoting the maximum search depth being considered. The Tanimoto kernel is defined by

\begin{equation}
    k_d^t(G_1,G_2) \defeq \frac{k_{\varphi_d}(G_1,G_2)}{k_{\varphi_d}(G_1,G_1) + k_{\varphi_d}(G_2,G_2) - k_{\varphi_d}(G_1,G_2)}.
\end{equation}
\end{definition}

The Tanimoto kernel essentially computes the ratio between the number of features extracted from both $G_1$ and $G_2$, and the total number of features extracted from $G_1$ and $G_2$ (without double counting). The codomain of the kernel is $[0,1]$.

\begin{definition}
Let $G_1,\,G_2$ be two graphs, and $d\in\mathbb{N}$. The MinMax kernel is defined as
\begin{equation}
    k_d^m(G_1,G_2) \defeq \frac{\sum_{p\in\mathcal{P}(d)} \min(\#\{ p \subset G_1 \},\,\#\{ p \subset G_2 \})}{\sum_{p\in\mathcal{P}(d)} \max(\#\{ p \subset G_1 \},\,\#\{ p \subset G_2 \})} 
\end{equation}

\noindent
where $\#\{p\subset G\}$ denotes the number of occurrences of the path p in G. 
\end{definition}

The biggest difference with the MinMax kernel is that a path appearing multiple times in a graph is factored into the calculation, where as the Tanimoto kernel only checks for one instance of a path. In the testing portion of the paper, this is the reason cited for why the MinMax kernel performs better on molecules with different sizes. 

\begin{definition}
\label{hybrid kernel definition}
Let $G_1,\,G_2$ be two graphs, $r\in\mathbb{N}$ representing feature vector length, and $d\in\mathbb{N}$ denoting the maximum search depth considered. Let $c\in(-1,2)$ be a user-defined parameter. The Hybrid kernel is defined as

\begin{equation}
    k_{d,r}^h(G_1,\,G_2) \defeq \frac{1}{3}[(2-c)(k_{d,r}^{t}(G_1,G_2)) + (1+c)(1-k_{d,r}^{t}(G_1,G_2))].
\end{equation}
\end{definition}

The notation $k_{d,r}^{t}$ is used to denote the Tanimoto kernel using the same base kernel as $k_{d,r}$. Taking one minus this altered Tanimoto kernel is equivalent to a logical negation. According to the authors of the paper, $c$ is often taken to be the average density of the feature vectors. This kernel is designed to be a convex combinations of two altered Tanimoto kernels.

\smallskip

Next, we discuss the complexity of these kernels following the analysis of \citet{ralaivola2005graph}. Using a suffix tree (see \citet{ukkonen1995line} for a construction algorithm) to store each path emanating from a particular node and not distinguishing orientation of paths, the complexity of each of the three kernels acting on two graphs $G_1$ and $G_2$ is $O(d n m)$, where $n$ and $m$ again represent the number of atoms and bonds of a graph, and $d$ is the maximum path-length in the search. Note that this can increase depending on the particulars of the depth-first search algorithm used.

\smallskip

\citet{ralaivola2005graph} tested their kernels using the voted perceptron classifier \citep{freund1999large} as a model on three data sets to test the prediction power of mutagenicity, toxicity, and anti-cancer activity. The first data set used was the MUTAG data set, the second was the PTC data set, and the third was the NCI GI50 data set. We were only able to find actives links to retrieve MUTAG. For kernel parameters, the depth was set to 10, $b$ was set to 1, and feature vectors of size of 512 and 1024 (when restricted) were both used. The tests showed that the Tanimoto and MinMax kernels in particular were quite effective in classification, with the MinMax kernel achieving the highest accuracy on the MUTAG data set at the time the paper was published. Both kernel were above 70\% accuracy on the NCI data set, with the MinMax kernel performing slightly better.

\smallskip

The Tanimoto kernel was also tested against the tree-pattern, marginalized, and extended marginalized graph kernels in the paper by \citet{sawada2014benchmarking}. A section of this paper is dedicated to benchmarking the predictive power of these four popular kernels on a data set of over 100,000 unique drug-target interactions using a chemogenomics approach. A pairwise kernel regression model was employed. The Tanimoto kernel performed best in many of the tests, and in the others remained very competitive. Detailed descriptions of the particular classification tasks can be found in the original paper.


\section{Boosting Kernel Performance}
\label{Boosting Kernel Performance section}

Many methods have been proposed to improve both classification accuracy and runtime performance of preexisting graph kernels. This can be done by extending existing kernels, for example the extension of the marginalized graph kernel discussed in section \ref{Marginalized Kernels}, or by introducing a general scheme that can be applied to classes of graph kernels. Our focus in this section will be on the latter.


\subsection{Graph RBF Kernel}
\label{Graph RBF Kernel Section}

While the RBF kernel (section \ref{Implementing Kernels in Nonlinear Classification section}) is commonly viewed as a standalone kernel with its own feature map and associated RKHS, one could also view it as a function composed with preexisting kernels. In its original formulation, it contains the squared Euclidean distance, which can be written as the sum of dot products, or \emph{linear kernels}. The use of feature maps is how the graph kernel variant of the RBF kernel is constructed. First, we present the \emph{kernel metric} defined in \citet{steinwart2008support}, which will replace the Euclidean metric for the general RBF kernel.

\begin{definition}
\label{kernel metric}
Let $k$ be the kernel of an RKHS on a set $U$ with feature map $\varphi$. The kernel metric on $U$ is defined as
\begin{equation}
    d_k(x,y) \defeq \| \varphi(x) - \varphi(y) \| = \sqrt{k(x,x) - 2k(x,y) + k(y,y)}
\end{equation}

\noindent
for every $x,\,y\in U$.
\end{definition}

\begin{remark}
The kernel metric is a pseudo-metric on $U$. For it to be a metric, we would also need for $\varphi$ to be injective.
\end{remark}

Now, if $k$ is some fixed graph kernel, then the \emph{graph RBF kernel composed with k} is defined as

\begin{equation}
    k_{\text{GRBF}} (G_1,G_2) \defeq \exp{\left(-\frac{d_k(G_1,G_2)^2}{2\sigma^2}\right)}
\end{equation}

\noindent
where $G_1$ and $G_2$ are two graphs, and $\sigma$ is a hyperparameter. The experiments conducted by \citet{kriege2020survey} used cross-validation to choose $\sigma$ from the set $\{ 2^{-7}, 2^{-6}, \cdots, 2^7 \}$. Overall, they found that the combination of the graph RBF kernel with other graph kernels led to a minor accuracy boost, typically a few percentage points. However, certain kernels received a large performance boost, such as the edge label histogram kernel (section \ref{simple examples of kernels section}), which became on-par with more complex kernels. The general trade-off is that optimizing $\sigma$ can be expensive, especially as the number of graphs in the data set grows. They recommend that the RBF kernel be used in conjunction with the vertex and edge label histogram kernels in any situation, and to avoid it with kernels such as the Weisfeiler-Lehman family of kernels who already have a hyperparameter to tune, and see negligible gains. \citet{sugiyama2015halting} composed the graph RBF kernel with the vertex edge label histogram kernel (definition \ref{vertex-edge label kernel definition}). They found a significant increase in classification accuracy with this variant of the kernel, with up to a 15\% increase compared to the original.


\subsection{Weisfeiler-Lehman Kernels}
\label{WL kernels}

Graphs \textemdash like most structures in mathematics \textemdash have a very important notion of \emph{equivalence} known as \emph{isomorphism}. Informally, two graphs $G_1$ and $G_2$ are said to be isomorphic if they have the same \emph{structure}. That is, if there is a bijection $\varphi: V_1 \mapsto V_2$ with the property that $(v,v')$ is an edge in $G_1$ if and only if $(\varphi(v),\varphi(v'))$ is an edge in $G_2$, and the corresponding vertices and edges with respect to $\varphi$ have the same labels (see \citet{kriege2020survey}, for example).

\smallskip

Weisfeiler-Lehman kernels (\citet{shervashidze2009fast}, \citet{shervashidze2011weisfeiler}) are based on a procedure known as the Weisfeiler-Lehman (WL) test of isomorphism \citep{weisfeiler1968reduction}. This is an algorithm that can be used to show that two graphs are \emph{not} isomorphic. No conclusion can be made if the algorithm terminates naturally. The idea is to iteratively assign labels to vertices of a graph depending on the previous labels of direct neighbours. These labels can then be compared between graphs; if they are the same, repeat the process, if not, the graphs are not isomorphic. The algorithm naturally terminates after $n$ iterations, where $n$ is the number of vertices in the graphs. The complexity of this algorithm is $O(hm)$, where $h$ is the number of iterations, which if leveraged by a graph kernel can offer much better computational-time scaling with respect to graph size than other popular kernels. Note that this complexity is dependent on the type of sorting algorithm implemented within the WL algorithm. The authors achieve this complexity through the implementation of a \emph{counting sort}. The full WL algorithm is presented in the paper by \citet{shervashidze2011weisfeiler}.

\smallskip

The general WL kernel is constructed as follows. For each iteration $i$ of the WL algorithm on a graph $G$, a new label $l_i$ for $G$ is created. Let $G_i = (V,E,l_i)$ be the \emph{WL graph} at height $i$. When $i=0$, we simply set $G_0=G$ and $l_0=l$. If $k$ is a graph kernel, then the \emph{WL kernel with h iterations and base kernel k} is defined as

\begin{equation}
\label{general WL kernel}
    K_{\textup{WL}}^h(G,G') \defeq \sum_{i=0}^h k(G_i,G_i') .
\end{equation}

This framework allows for many new kernels to leverage the label refinement produced by the WL algorithm. The first particular instance presented was introduced by \citet{shervashidze2009fast}, and is called the Weisfeiler-Lehman \emph{subtree kernel}. It was later shown by \citet{shervashidze2011weisfeiler} to be a special case of equation \ref{general WL kernel}. The latter formulation is what we will present. For context, the goal outlined in the original paper by \citet{shervashidze2009fast} was to develop a fast kernel on labeled graphs utilizing subtree structure. As the authors note, methods for speeding up computational time for walk-based and subgraph-based kernels were known, but the same could not be said for graph kernels using trees. It is known that graph kernels scale poorly with respect to graph size, which is a limiting factor on any method. For two graphs $G_1$ and $G_2$, the \emph{Weisfeiler-Lehman subtree kernel with h iterations}, denoted by $k_{WLS}^h$, is given by the general WL kernel in equation \ref{general WL kernel} with the vertex label kernel as its base (section \ref{simple examples of kernels section}). Hence, it can be written as 

\begin{equation}
    k_{\textup{WLS}}^h (G_1,G_2) = \sum_{i=0}^h \sum_{v_1\in V_1} \sum_{v_2\in V_2} \textbf{I}(l_i(v_1)=l_i(v_2)).
\end{equation}

Note that the value of this kernel is a byproduct of running the algorithm on two graphs. This implies that complexity of this kernel is $O(hm)$ \citep[Theorem 5]{shervashidze2011weisfeiler}. However, the author provide an algorithm based on explicit feature map computation that allows for the Gram matrix of $k_{WLS}^h$ on $N$ graphs to be computed in time $O(Nhm + N^2hn)$, where $m$ and $n$ are the maximum number of edges and vertices (respectively) over the $N$ graphs. See section \ref{Notes on explicit and implicit computation section} for a short discussion regarding efficiency gains from explicit feature map computation.

\smallskip

The next formulation in \cite{shervashidze2011weisfeiler} is the \emph{WL edge kernel}, which is again given by the general framework in equation \ref{general WL kernel}, but now with the edge label kernel as its base (section \ref{simple examples of kernels section}). As the base kernel has complexity $O(m^2)$, an upper bound on the complexity of the WL edge kernel is $O(hm^2)$. The final WL kernel discussed employs the shortest-path kernel introduced by \citet{borgwardt2005shortest} for the base kernel (see section \ref{Shortest-path Graph Kernel section}). We briefly review concepts from that section. The first step in computing the WL-SP kernel is to determine the shortest-paths between any two vertices of a graph. In \cite{borgwardt2005shortest}, this is done by applying the Floyd-transformation to a graph, which outputs a matrix whose $ij$th entry is the distance (measured in terms of number of edges for chemical graphs) between vertices $v_i$ and $v_j$. The shortest-path kernel $k_{sp}$ then counts how many shortest-length paths in each graph share the same path length and the same labels for the start and end vertices. The complexity for computing this kernel has an upper bound of $O(hn^4 + hm)$.

\smallskip

\citet{shervashidze2011weisfeiler} conducted two sets of experiments. The first is an empirical assessment of the runtime performance of the WL subtree kernel on toy data sets using both the pairwise kernel computation scheme, as well as the ``global" $N$-graph implementation. The results showed that ``global" option was far quicker, and appears to scale well with graph size. The second compares the predicting power of three iterations of WL kernels against popular graph kernel of the time using standard benchmarking data sets. Many kernels were used in this experiment, including the generalized random-walk kernel from \cite{vishwanathan2010graph}, the graphlet kernel of \cite{shervashidze2009efficient}. LIBSVM was implemented for SVM classification, and experiments were repeated 10 times. The height $h$ for the WL subtree kernel was chosen via 10-fold cross-validation in the range $\{ 0,1,\,\cdots,10 \}$. Note that for WL edge and WL shortest-path kernels, values of 2 and 3 for $h$ were chosen by the model almost exclusively. The results showed that the WL subtree kernel was able to handle large graphs with thousands of vertices, and was very competitive on small data sets too. For the D\&D data set containing 1178 protein graphs, the WL kernel took 11 minutes to compute, where as the other WL-based kernels took anywhere between 23 hours and over a year. In terms of classification performance, the WL subtree kernel did particularly well on the NCI1 and NCI109 data sets. The WL shortest-path kernel performed far better than the other kernels on the smaller ENZYMES data set. Overall, the WL kernels were either outperforming other kernels, or about on-par with them, in terms of both classification accuracy and CPU runtime.

\smallskip

\citet{kriege2016valid} used the WL kernel framework to develop the WL optimal assignment kernel (section \ref{Optimal assignment kernels section}), which according to the tests done in \citet{kriege2020survey}, performs better than the standard WL subtree kernel. \citet{morris2021power} discusses the applications of the WL algorithm in machine learning, including the WL kernel and beyond.


\subsection{Optimal Assignment Kernels}
\label{Optimal assignment kernels section}

So far, all of the graph kernels we have discussed have been instances of the convolution kernel framework. In this section, we briefly discuss a newer class of kernels called \emph{optimal assignment kernels} that leverage a different framework. While convolution kernels sum over all possible substructures of a certain type, optimal assignment kernels work by selecting the particular pairings of substructures that maximized the function value. Many variations of the optimal assignment kernel have appeared in the literature. After being introduced by \citet{frohlich2005optimal}, it was pointed out by \citet{rupp2007kernel} that the kernels need not be positive definite. The loss of this property impacts the functionality when using support vector machines as positive definiteness is needed to guarantee that the corresponding optimization problem has a global minimum \citep{rupp2007kernel}. \cite{rupp2007kernel} proposed a modification to the ``kernel" by \citet{frohlich2005optimal} which is sometimes called the \emph{iterative similarity optimal assignment kernel} (ISOAK) \citep[pg.\ 217]{dehmer2012statistical}. They used it in place of a kernel in an SVM to gather experimental results and provide empirical evidence for instances where the measure is indeed positive definite. We will focus on the more recent work done by \citet{kriege2016valid}, which provided sufficient conditions for the base kernel that guarantee that the optimal assignment kernel is positive definite. This was the first successful attempt at an optimal assignment kernel framework. Note that their presentation is far more general than what is found in this section. Our goal ultimately will be to understand optimal assignment kernels in conjunction with the Weisfeiler-Lehman algorithm. First, we present a formal definition of the optimal assignment kernel from \citet{kriege2020survey}.

\begin{definition}
Let $(x_1,\, \cdots,\, x_n)$, $(y_1,\, \cdots,\, y_n)$ be decomposition's of graphs $G_1$, $G_2$ (respectfully) into subgraphs, and let $k$ be a base kernel defined on those subgraphs. The optimal assignment kernel on $G_1$ and $G_2$ is

\begin{equation}
    K_{\textup{OA}}(G_1,\,G_2) = \max_{\pi\in\Pi_n} \sum_{i=1}^n k(x_i, y_{\pi(i)}) 
\end{equation}

\noindent
where $\Pi_n$ is the set of permutations of the set $\{ 1,\,\cdots,\,n\}$.
\end{definition}

\smallskip

Next, we present the key definition from \citet{kriege2016valid}.

\begin{definition}
\label{strong kernel definition}
    Let X be a nonempty set. A kernel $k: X\times X \mapsto [0,\infty)$ is said to be strong if for any $v,\, u,\, w \in X$, we have $k(v,u) \geq \min(k(v,w),\,k(w,u))$.
\end{definition}

\citet{kriege2016valid} proved that the optimal assignment similarity measure is positive definite if its base kernel is strong. As a very important special case, we now discuss the Weisfeiler-Lehman (WL) implementation of the optimal assignment kernel. Let $G$ be a graph. At the beginning of the WL algorithm, every vertex in $G$ is given the same label. During each iteration, the vertices of $G$ are relabeled (referred to as \emph{colours} by the authors) based on how many neighbours they have. Note that the relabeling function must be injective. We can relate vertices using a tree $T$ whose nodes contain the colours used in the algorithm. During the \emph{i}th recolouring, leaf nodes are added to $T$ that represent the current colours of the graph. The parent node of a given leaf node contains the previous colouring of the associated graph vertices. After the last ($h$th) iteration, we can define the feature map $\varphi$ of the kernel acting on the vertex labels (colours) of the final leaf nodes. Let $c$ denote the number of colours used during the WL algorithm. If $v$ is a vertex of the graph $G$, then $\varphi(v)$ is a $n$-dimensional vector whose \emph{j}th coordinate is 1 if the \emph{j}th node appears in the path between the root node and the leaf node associated with $v$, and 0 otherwise. Essentially, this feature map encodes the colour history of each vertex in $G$. See \citet[Figure 3.11]{borgwardt2020graph} for an illustration on how this algorithm works. The \emph{histogram} of the graph $G$ after $h$ iterations is then defined as

\begin{equation}
    H^h(G) = \sum_{v\in V} \varphi(v).
\end{equation}

This representation of graphs is used in the equivalent form of the optimal assignment kernel \citep[Theorem 2]{kriege2016valid}. This form is what is presented in equation \ref{wloa kernel}. First, we need one final definition from \citet{swain1991color}.


\begin{definition}
The histogram intersection kernel defined on two vectors $\textup{\textbf{x}},\, \textup{\textbf{y}} \in \mathbb{R}^n$ is defined as
    \begin{equation}
        k_{\cap}(\textup{\textbf{x}},\textup{\textbf{y}}) \defeq \sum_{i=1}^n \min(x_i,y_i) .
    \end{equation}
\end{definition}

This was shown to be a kernel on the set of binary vectors by \citet{barla2003histogram}. The proof for the generalization to all of $\mathbb{R}^n$ is found in \citet{boughorbel2005generalized}. Now, let $G_1$ and $G_2$ be two graphs. Using \citep[Theorem 2]{kriege2016valid}, the Weisfeiler-Lehman optimal assignment (WL-OA) kernel with $h$ iterations on $G_1$ and $G_2$ can be written as

\begin{equation}
\label{wloa kernel}
        k_{\textup{WLOA}}^{h} (G_1, G_2) = k_{\cap} (H^h(G_1),H^h(G_2))).
\end{equation}

Optimal assignment versions of the vertex and edge label kernels are also described in \cite{kriege2016valid}. Due to the histogram intersection, the complexity of computing the WL-OA kernel is the same as the complexity of computing the WL algorithm, which is $O(nh)$, where $n$ is the number of vertices in the graph \citep{borgwardt2020graph}. This was also empirically observed during the experiments in \citet{kriege2016valid}, which we discuss next.

\smallskip

In \citet{kriege2016valid}, experiments were conducted on a few implementations of optimal assignment kernels, including WL-OA, as well as their non-optimal assignment counterparts. The authors used support vector machines for classification, implemented with LIBSVM. The experiments were conducted using 10-fold cross validation, and were repeated 10 times. The number of iterations $h$ for the WL algorithm were taken between 0 and 7. Many of the common benchmark data sets were used, including the MUTAG, PTC-MR, NCI1, and NCI109 data sets containing small molecules. For testing with large molecules, they used PROTEINS, D\&D, and ENZYMES. Finally, they used the COLLAB and REDDIT data sets, which contain social network data. These data sets are freely available from \href{https://chrsmrrs.github.io/datasets/}{TUDataset}. Other than the MUTAG data set, the optimal assignment kernel implementations consistently outperformed the convolution implementations, in some cases by 10\%. The WL-OA was the best performing kernel on 7/9 data sets, while still being very competitive on the remaining 2. As expected due to its complexity, the WL-OA kernel computation was very fast. In another experiment by \citet{kriege2020survey}, a large number of common kernels were tested, including WLOA. It was concluded that both the WL and WL-OA kernels were the most accurate for a majority of the data sets (although nearly every kernel was best on at least one). The authors recommended implementing WL-OA when dealing with small-to-medium-sized data sets and using an SVM for classification.


\subsection{Notes on Explicit and Implicit Computation}
\label{Notes on explicit and implicit computation section}

As mentioned in chapter \ref{chapter2}, one of the benefits of using kernels is that the rich structure of the associated RKHS can be accessed without needing to compute the feature map, or even know its explicit form. However, as noted by \citet[pg.\ 24]{kriege2020survey}, some common implementations of graph kernels transform graph data into feature vectors directly and then compute the inner product between the representations of the graphs. \cite{kriege2014explicit} investigated whether the kernel trick is beneficial for graph kernels by computing both explicit (feature map and dot product) and implicit maps, and comparing the runtime performance between each computation. In some instances, they observed that it is faster to store graphs in feature vectors and compute dot products, rather than use implicit evaluation. For example, walk-based kernels working with a small walk-length have superior runtime performance when evaluated explicitly. Once the walk-length gets large, the runtime for explicit evaluation increases drastically, surpassing the implicit computation time that increased linearly throughout the entire experiment. Algorithms for explicit and implicit of the \emph{k}-walk kernel computations are given by \cite{kriege2014explicit}. In the end, empirical evidence showed that kernel computations in the context of chemoinformatic applications may see improvements in runtime in some situations when solely using explicit kernel representations, or adapting a hybrid model where a switch between explicit and implicit occurs after a certain predetermined point, such as a walk length or number of iterations of the WL algorithm. This is suggested by the evidence that a small number of labels benefits from explicit computation, as well as shorter walk-lengths for walk-based kernels. When using Weisfeiler-Lehman label refinement on the ENZYMES data set, the results suggested that the best performance is obtained when the explicit computation is used initially (i.e.\ before refinement), and the implicit computation is used for every iteration of the label refinement.


\section{Software and Data sets}
\label{Software and Data sets section}

This section provides short descriptions of the data sets and software commonly found throughout graph kernel literature, as well as download links when available. 


\subsection{Common Data Sets}
\label{common data sets section}

\smallskip

The following is a description of the most important data sets mentioned throughout the chapter, directly taken from the paper by \citet{shervashidze2011weisfeiler}, unless otherwise stated. \emph{MUTAG} \citep{debnath1991structure} is a data set of 188 mutagenic aromatic and heteroaromatic nitro compounds labeled according to whether or not they have a mutagenic effect on the Gram-negative bacterium Salmonella typhimurium. They are split into two classes: 125 positive examples with high mutagenic activity (positive levels of log mutagenicity), and 63 negative examples with no or low mutagenic activity, and they are made of 26 atoms and 27.9 covalent bonds in average (Direct quote from \citet{mahe2009graph}). \emph{NCI1} and \emph{NCI109} represent two balanced subsets of data sets of chemical compounds screened for activity against non-small cell lung cancer and ovarian cancer cell lines, respectively (\citet{wale2008comparison}, and \href{http://pubchem.ncbi.nlm.nih.gov}{\color{blue}{PubChem}}). \emph{ENZYMES} is a data set of protein tertiary structures obtained from \citet{borgwardt2005protein} consisting of 600 enzymes from the BRENDA enzyme database \citep{schomburg2004brenda}. In this case the task is to correctly assign each enzyme to one of the 6 EC top-level classes. \emph{D\,\&D} is a data set of 1178 protein structures \citep{dobson2003distinguishing}. Each protein is represented by a graph, in which the nodes are amino acids and two nodes are connected by an edge if they are less than 6 Angstroms apart. The prediction task is to classify the protein structures into enzymes and non-enzymes. The Predictive Toxicology Challenge (PTC) data set \citep{helma2001predictive} reports the carcinogenicity of several hundred chemical compounds for Male Mice (MM), Female Mice (FM), Male Rats (MR) and Female Rats (FR) (Direct quote from \citet{ralaivola2005graph}). 

\smallskip

\href{https://chrsmrrs.github.io/datasets/}{\color{blue}{TUDataset}} is repository containing many benchmarking data sets, including MUTAG, PTC-MR, NCI1, NCI109, PROTEINS, D\&D, and ENZYMES. It also contains data set for social network graphs, such as COLLAB and REDDIT, mentioned in section \ref{Optimal assignment kernels section}.


\subsection{Computing Graph Kernels}
\label{Kernel computation section}

\href{https://github.com/BorgwardtLab/graph-kernels}{\color{blue}{Graph-kernels}} is a GitHub repository containing both \textsf{R} and \texttt{C++} code to compute vertex/edge label kernels, common random walk kernels, the Weisfeiler-Lehman graph kernel, as well as many others. The documentation can be found \href{https://cran.r-project.org/web/packages/graphkernels/graphkernels.pdf}{\color{blue}{here}}. For a Python version, see \href{https://github.com/BorgwardtLab/GraphKernels}{\color{blue}{this page}}. \href{https://bsse.ethz.ch/mlcb/research/machine-learning/graph-kernels.html}{\color{blue}{ETH-Zurich}} has a link to the graph-kernels package, as well as code for other graph kernel implementations and common chemoinformatic data. \href{https://www.csie.ntu.edu.tw/~cjlin/libsvm/}{\color{blue}{LIBSVM}} is a popular library for implementing support vector machines, and has been ported to many languages. \href{http://chemcpp.sourceforge.net/html/index.html}{\color{blue}{ChemCPP}} is a \texttt{C++} library containing functions to compute graph kernels on chemical compounds. The kernels available include the marginalizes graph kernels and extensions (section \ref{Marginalized Kernels}), Tree-pattern kernels (section \ref{Tree-Pattern Graph Kernel section}), and Tanimoto kernels (\ref{Tanimoto Kernels Section}). The \href{http://pyml.sourceforge.net/}{\color{blue}{PyML}} framework has tools to perform machine learning techniques in python such as support vector classification, and is compatible with Linux and macOS. \href{https://svm.msl.ubc.ca/gist/}{\color{blue}{GIST}} is another alternative for support vector machine classification in the \texttt{C} programming language. \href{https://mrupp.info/publications.html}{Matthias Rupp's} personal web page contains a download for a Java implementation of the ISOAK kernel (section \ref{Optimal assignment kernels section}).


\section{Further Reading}
\label{Further Reading section}

After nearly 20 years of advances, the graph kernel literature has grown enormously. The goal of this chapter was to give the reader a broad overview of different families of graph kernels, as well as insight into their performance. Each paper sighted throughout can be studied for many more technical details. In this section, we highlight some of the key sources to learn more about graph kernels in a general setting, as well as particular graph kernels that were not examined, but may be of interest.

\smallskip

While typical graph kernel sources are that of conference proceedings and published papers, the textbook \citet[pg.\,217]{dehmer2012statistical} contains a chapter on graph kernels, and serves as an excellent introduction to the topic and the literature. It also contains a section with references to bio and chemoinformatic applications of graph kernels. Various published reviews of the graph kernel literature exist as well. \citet{kriege2020survey} conduct an experimental study on many different types of graph kernels and data sets relevant to many fields. The paper also provides a general guide on how to choose kernels based on properties of the graphs such as size, structure, and labeling. In \citet{borgwardt2020graph}, an overview of the many families of graph kernels is provided, along with large-scale comparisons between kernels on many standard data sets, and a discussion on where the field is headed. \citet{ghosh2018journey} give a technical overview of a wide-variety of graph kernels, including many ``modern" options, as well as experimental results on many types of data sets.

\smallskip

\citet{shervashidze2009efficient} introduces two graph kernels that count subgraphs with a set number of vertices, called \emph{Graphlets}, with the goal of efficiently handling large graphs. \citet{togninalli2019wasserstein} extend the Weisfeiler-Lehman graph kernel to graphs with weighted-edges and continuous vertex attributes using Wasserstein distance. \citet{gauzere2012two} introduce two kernels, the \emph{Treelet} and \emph{Graph Laplacian} kernel, with the goal of direct application to problems in chemoinformatics. \citet{vishwanathan2010graph} define the Composite graph kernel, which is the sum of a kernel on a pair of graph and the same kernel on the respective complement graphs. The hope is that this modification may improve performance in situations where the absence of interaction between atoms is also important, such as in protein interaction in disease \citep{vishwanathan2010graph}. \citet{mahe2006pharmacophore} introduce the \emph{Pharmacophore kernel} that acts on 3D representations of molecules. Pharmacophore kernels are in this sense not graph kernels themselves, however they still are positive-definite kernels. The authors also show that this kernel in a sense extends the random walk graph kernels of \citet{gartner2003graph} (section \ref{Direct Product Kernel section}).

\nocite{*}
\sloppy 
\bibliography{main.bib}
\end{document}